%% file: main.tex
\documentclass[twoside]{article}

\usepackage[accepted]{aistats2021}

\usepackage[utf8]{inputenc} 
\usepackage[T1]{fontenc}    
\usepackage{hyperref}       
\usepackage{url}            
\usepackage{booktabs}       
\usepackage{amsfonts}       
\usepackage{nicefrac}       
\usepackage{microtype}      

\usepackage{amsmath}

\usepackage{natbib}
\usepackage[ruled,vlined]{algorithm2e}

\usepackage{amsmath}
\usepackage{tikz}
\usepackage{mathdots}
\usepackage{yhmath}
\usepackage{cancel}
\usepackage{mathtools}
\usepackage{color}
\usepackage{siunitx}
\usepackage{array}
\usepackage{multirow}
\usepackage{amssymb}
\usepackage{gensymb}
\usepackage{tabularx}
\usepackage{booktabs}
\usepackage{bbm, dsfont}
\usepackage{amsthm}
\usepackage{bigints}
\usepackage{float}
\DeclarePairedDelimiter\ceil{\lceil}{\rceil}
\DeclarePairedDelimiter\floor{\lfloor}{\rfloor}
\usetikzlibrary{fadings}
\usetikzlibrary{patterns}
\usetikzlibrary{shadows.blur}
\usepackage{hyperref}
\usepackage{esvect}
\usepackage{algorithm2e}
\allowdisplaybreaks
\usepackage[symbol]{footmisc}

\usepackage[frozencache,cachedir=.]{minted}
\usepackage{color}
\usepackage{xcolor}

\newcommand{\removelatexerror}{\let\@latex@error\@gobble}

\input{math_commands}

\begin{document}
\twocolumn[

\aistatstitle{Representation of Reinforcement Learning Policies in Reproducing Kernel Hilbert Spaces}

\aistatsauthor{ Bogdan Mazoure$^{*}$\footnotemark[1]\footnotemark[2] \And Thang Doan$^{*}$\footnotemark[1]\footnotemark[2] \And  Tianyu Li\footnotemark[1]\footnotemark[2] \AND Vladimir Makarenkov\footnotemark[3] \And Joelle Pineau\footnotemark[1]\footnotemark[2]\footnotemark[4]\footnotemark[5] \And Doina Precup\footnotemark[1]\footnotemark[2]\footnotemark[5]\footnotemark[6] \And Guillaume Rabusseau\footnotemark[2]\footnotemark[5]\footnotemark[7]}

\aistatsaddress{ } 
]
 
\begin{abstract}
We propose a general framework for policy representation for reinforcement learning tasks. This framework involves finding a low-dimensional embedding of the policy on a reproducing kernel Hilbert space (RKHS). The usage of RKHS based methods allows us to derive strong theoretical guarantees on the expected return of the reconstructed policy. Such guarantees are typically lacking in black-box models, but are very desirable in tasks requiring stability. We conduct several experiments on classic RL domains. The results confirm that the policies can be robustly embedded in a low-dimensional space while the embedded policy incurs almost no decrease in return.
\end{abstract}

\footnotetext[1]{McGill University    \footnotemark[2] Mila - Quebec Artificial Intelligence Institute \footnotemark[3] Universit\'e du Qu\'ebec \`a Montr\'eal \footnotemark[4] Facebook AI Research \footnotemark[5] CIFAR AI chair \footnotemark[6] DeepMind \footnotemark[7] Université de Montréal. $^{*}$Equal contribution. Correspondence to: Bogdan Mazoure <bogdan.mazoure@mail.mcgill.ca>}

\section{Introduction}
\input{introduction.tex}

\section{Preliminary}
In this section, we provide a brief introduction to reinforcement learning, density approximation, as well as singular value decomposition.
\input{preliminary.tex}

\section{Framework}
\input{theory.tex}

\section{Practical algorithm}
\input{algorithm.tex}

\section{Experimental results}
\input{experiments.tex}

\section{Conclusion}
\input{conclusion.tex}

\section*{Acknowledgements}
This research is supported by the Canadian Institute for Advanced Research (CIFAR AI chair program). 


\input{main.bbl}
\clearpage
\newpage
\onecolumn
\section{Appendix}
\input{appendix.tex}
\end{document}

%% file: math_commands.tex
\usepackage{amsmath,amsfonts,bm,bbm}

\def\eqref#1{equation~\ref{#1}}

\def\ceil#1{\lceil #1 \rceil}
\def\floor#1{\lfloor #1 \rfloor}
\def\1{\bm{1}}

\DeclareMathAlphabet{\mathsfit}{\encodingdefault}{\sfdefault}{m}{sl}
\SetMathAlphabet{\mathsfit}{bold}{\encodingdefault}{\sfdefault}{bx}{n}

\newcommand{\cA}{\mathcal{A}}

\newcommand{\cH}{\mathcal{H}}

\newcommand{\cS}{\mathcal{S}}

\newtheorem{theorem}{Theorem}[section]
\newtheorem{corollary}{Corollary}[theorem]
\newtheorem{lemma}[theorem]{Lemma}

\renewcommand{\vec}[1]{\ensuremath{\bm{#1}}}

\newcommand{\mat}[1]{\ensuremath{\mathbf{#1}}}

%% file: introduction.tex
In the reinforcement learning (RL) framework, the goal of a rational agent consists in maximizing the expected rewards in a dynamical system by finding a suitable conditional distribution known as \textit{policy}. This conditional distribution can be found using policy iteration and any suitable function approximator, ranging from linear models to neural networks~\citep{sutton2018reinforcement,ddpg}. Albeit neural networks being the state-of-the-art learner for most performance-based tasks~\citep{trpo,bellemare2017distributional}, this class of blackbox models provide few guarantees on collected rewards, which should be imposed on a policy in risk-sensitive tasks~\citep{garcia2015comprehensive}. 


In many application domains such as self-driving cars or robotic controllers~\citep{sadigh2016information,akametalu2014reachability,berkenkamp2017safe,katz2017reluplex}, it is often crucial to have a robust policy. One criteria of such policy is that the variance of the expected returns should be as small as possible. Unfortunately, due to the non-convexity of most deep learning methods, it is difficult to compute the exact variance. Moreover, the high complexity of these models often leads to high variance and has been shown empirically~\citep{pinto2017robust}.

One approach to mitigate this issue is to represent complex policies using decision rules~\citep{bastani2018verifiable,katz2017reluplex,aswani2013provably}. However, their major downsides are non-convex loss functions (preventing in-depth error analysis), and access to external oracle information (e.g. unbiased gradients).

In order to address the need for performance guarantees, we propose a principled way to represent a broad class of policy density functions as a point in a Reproducing Kernel Hilbert Space (RKHS). One advantage of such method is that it is straight-forward to truncate over the RKHS, which leads to a reduction in variance of the expected returns. In addition, the theoretical framework of RKHS offers neat tools for analyzing the error introduced by the truncation. In general, our framework achieves two desirable properties: (i) our method tends to produce policies with lower return variance than in the original policy; (ii) the embedding dimensionality of the studied policies can vary without large performance drops. We show both of these properties theoretically as well as empirically.

The use of RKHS and kernel methods in reinforcement learning problems is not uncommon. Non-parametric kernel density estimators have previously been used to represent fundamental RL quantities such as the transition dynamics~\citep{grunewalder2012modelling,nishiyama2012hilbert,lever2016compressed,barreto2016practical} and predictive representations of states~\citep{littman2002predictive} known as PSRs. Moreover, some works leverage spectral learning and RKHS to extend the classical PSR setting~\citep{boots2013hilbert,li2020efficient}. One can even recover the original distribution from the kernel mean estimator through the tools provided by \cite{kanagawa2014recovering}, which turns out to be useful in message passing \citep{song2008tailoring}.

Our contributions are the following:
\begin{itemize}
    \item We propose a decomposition of policy densities within a separable Hilbert space and derive a set of performance bounds which, when used together, provide a heuristic to pick the dimensionality of the embedding.
    \item The truncation of the RKHS embedding is shown to reduce variance of returns with respect to the initial conditions, when second order moment conditions are satistfied.
    \item We interpret the performance error bounds through three separate terms: truncation (Thm.~\ref{thm:mdp_rewards_trpo}), discretization (Thm.~\ref{thm:discretization_thm}), and pruning (Thm.~\ref{thm:pruning}).
    \item Empirical evidence on continuous control tasks supports our theoretical findings.
\end{itemize}
  To the best of our knowledge, this is the first work proposing a general framework to learn an RKHS policy embedding with performance and long-term behaviour guarantees. 

%% file: preliminary.tex
\subsection{Reinforcement learning}
We consider a problem modeled by a discounted Markov Decision Process (MDP) 
$\mathcal{M}:=
\left\langle
    \mathcal{S},
    \mathcal{A},
    r,
    P,
    S_0,
    \gamma
\right\rangle
$,
where 
$\mathcal{S}$ is the state space;
$\mathcal{A}$ is the action space; given states $s$, $s'\in\mathcal{S}$, action $a\in\mathcal{A}$,
$\mathbb{P}(s'|s, a)$ is the transition probability of transferring $s$ to $s'$ under the action $a$ and
$r(s, a)$ is the reward collected at state $s$ after executing the action $a$;
$S_0\subseteq \cS$ is the set of initial states and
$\gamma$ is the discount factor. Through the paper, we assume that $r$ is a bounded function.

The agent in RL environment often execute actions according to some policies. We define a stochastic policy $\pi: \mathcal{S}\times \mathcal{A} \rightarrow [0, 1]$ such that $\pi(a|s)$ is the conditional probability that the agent takes action $a\in\mathcal{A}$ after observing the state $s\in\mathcal{S}$.
The objective is to find a policy $\pi$ which maximizes the expected discounted reward:
\begin{align}
    \eta(\pi)=\mathbb{E}_{\pi}\left[\sum_{t=0}^\infty\gamma^tr(s_t, a_t)\right].
    \label{eq:eta_mdp}
\end{align}
The state value function $V_\pi$ and the state-action value function $Q_\pi$ are defined as:
\begin{align*}
    V_{\pi}(s)&=\mathbb{E}_{\pi}\left[\sum_{k=t}^\infty\gamma^{k-t}r(s_k, a_k)\bigg|s_t=s\right],\\
    Q_{\pi}(s, a)&=\mathbb{E}_{\pi}\left[\sum_{k=t}^\infty\gamma^{k-t}r(s_k, a_k)\bigg|s_t=s,a_t=a\right]
\end{align*}
The gap between $Q_\pi$ and $V_\pi$ is known as the advantage function:
 \begin{align*}
     A_{\pi}(s,a)&=Q_{\pi}(s, a)-V_{\pi}(s),
 \end{align*}
 where $a \sim \pi(\cdot|s)$. The coverage of policy $\pi$ is defined as the stationary state visitation probability $\rho_\pi(s)=\mathbb{P}(s_t=s)$ under $\pi$.  

\subsection{Function approximation in inner product spaces}
The theory of inner product spaces has been widely used to characterize approximate learning of Markov decision processes~\citep{puterman2014markov,tsitsiklis1999optimal}. In this subsection, we provide a short overview of inner product spaces of square-integrable functions on a closed interval $I$ denoted $L^2(I)$.

The $L^2(I)$ space is known to be separable for $I=[0,1]$, i.e., it admits a countable orthonormal basis of functions $\{\omega_k\}_{k=1}^\infty$, such that $\langle \omega_i,\omega_j\rangle = \delta_{ij}$ for all $i,j$ and $\delta$ being Kronecker's delta. This property makes possible computations with respect to the inner product
\begin{equation}
    \langle f,g \rangle = \int_{t\in I} f(t)\bar{g}(t)dt,
\end{equation}
and corresponding norm $||f||=\sqrt{\langle f, f \rangle}$, for $f,g\in L^2(I)$.

That is, for any $f\in L^2(I)$, there exist scalars $\{\xi^f_k\}_{k=1}^\infty$ such that its representation in the RKHS
\begin{equation}
    \hat{f}(t)=\sum_{k=1}^\infty \xi^f_k \omega_k(t),\;\; \xi^f_k=\langle f,\omega_k\rangle\;.
    \label{eq:linear_comb_L2}
\end{equation}

 If $\hat{f}_{K}(t)=\sum_{k=1}^K\xi^f_k \omega_k(t)$, then adding $\xi_{K+1}\omega_{K+1}$ to $\hat{f}_{K}$ can be thought of as adding a new orthogonal basis.

Eq.~\eqref{eq:linear_comb_L2} suggests a simple embedding rule, where one would project a density function onto a fixed basis and store only the first $\{\xi^f_k\}_{k=1}^K$ coefficients. Which components to pick will depend on the nature of the basis: harmonic and wavelet bases are ranked according to amplitude, while singular vectors are sorted by corresponding singular values.

The choice of the basis $\{\omega_k\}_{k=1}^\infty$ plays a crucial role in the quality of approximation. The properties of the function $f$ (e.g. periodicity, continuity) should also be taken taken into account when picking $\omega$. Some examples of well-known orthonormal bases of $L^2(I)$ are \emph{Fourier} $\omega_k(t)=e^{2\pi i k t}$~\citep{fourier_analysis_book} and \emph{Haar} $\omega_{k,k'}(t)=2^{k/2}\omega(2^kt-k')$~\citep{haar1909theorie}. Other examples include but are not limited to Hermite and Daubechies bases~\citep{daubechies1988orthonormal}. Moreover, it is known that a mixture model with countable number of components is a universal density approximator \citep{mclachlan2004finite}; in such case, the basis consists of Gaussian density functions and is not necessarily orthonormal.

Moreover, matrix decomposition algorithms such as the truncated singular value decomposition (SVD) are popular projection methods due to their robustness and extension to finite-rank bounded operators~\citep{zhu2007operator}, since they learn both the basis vectors and the corresponding linear weights. Although SVD has previously been used in embedding schemes~\citep{lu2017fully,goetschalckx2018efficiently}, unlike us, most works apply it on the weights of the function rather than the outputs. Similarly to the density approximators above, a key appeal of SVD is that the error rate of truncation can be controlled efficiently through the magnitude of the minimum truncated singular values\footnote{see Eckart-Young-Mirsky theorem}.

While convergence guarantees are mostly known for the closed interval $I=[0,1]$, multiple works have studied properties of the previously discussed basis functions on the whole real line and in multiple dimensions~\citep{egozcue2006hilbert}. Weaker convergence results in Hilbert spaces can be stated with respect to the space's inner product. If, for every $g\in L^2(I)$ and sequence of functions $f_1,..,f_n\in L^2(I)$,
\begin{equation}
    \lim_{n\to \infty}\langle f_n,g\rangle \to \langle f,g\rangle,
\end{equation}
then the sequence $f_n$ is said to converge weakly to $f$ as $n\to \infty$. Stronger theorems are known for specific bases but require stronger assumptions on $f$.

In order to allow our learning framework to have strong convergence results in the inner product sense as well as a countable set of basis functions, we restrict ourselves to separable Hilbert spaces.

%% file: theory.tex
  
In this section, we first introduce a framework to represent policies in an RKHS with desirable error guarantees. The size of this embedding can be varied through the coefficient truncation step.
We show that this truncation has the property to reduce the return variance without drastically affecting the expected return. 
Next, we propose a discretization and pruning steps as a tractable alternative to working directly in the continuous space. Finally, we derive a practical algorithm which leverages all aforementioned steps.

\subsection{Formulation as RKHS problem}
By Mercer's theorem~\citep{mercer1909functions}, any symmetric positive-definite kernel function $\kappa$ with associated integral operator $T_\kappa$ can be decomposed as:
\begin{equation}
\begin{split}
    \kappa(x,y)&=\sum_{k=1}^\infty (\sqrt{\lambda_k}e_k(x))(\sqrt{\lambda_k}e_k(y))\\
    &=\sum_{k=1}^\infty \omega_k(x)\omega_k(y),\\
\end{split}
\end{equation}
where $\lambda_k$ and $e_k$ are the eigenvalues and eigenfunctions of $T_\kappa$, respectively.

It follows by Moore–Aronszajn theorem~\citep{aronszajn1950theory} that there exists a unique Hilbert space $\cH$ of functions for which $\kappa$ is the kernel.

Moreover, the inner product associated with $\cH$ is:
\begin{equation}
    \langle f,g \rangle_\cH := \sum_{k=1}^\infty \frac{\langle f,e_k\rangle_{L_2} \langle g, e_k\rangle_{L_2}}{\lambda_k}=\sum_{k=1}^\infty \frac{\xi^f_k \xi_k^g}{\lambda_k}.
\end{equation}
This allows us to state performance bounds between policies embedded in an RKHS in terms of their projection weights, as shown in the next section.

Consider the conditional distribution $\pi:\cS\times \cA \to \mathbb{R}$. For a fixed $s$, the function $\pi(\cdot|s)$ belongs to a Hilbert space $\cH_s$ of functions $f:\cA\to\mathbb{R}$.
  
\begin{lemma}
Let $s$ be a state in $\cS$. Let $\cH_s$ be an RKHS, the integral operator of which has eigenvalues $\{\lambda_k\}_{k=1}^\infty$ and eigenfunctions $\{e_k\}_{k=1}^\infty$. Let $\xi_k^{\pi_i}=\langle \pi_i,e_k\rangle$ and  $\pi_{1},\pi_{2}\in \cH_s$ such that $\pi_1=\pi_1(\cdot|s),\pi_2=\pi_2(\cdot|s)$. Then, the following holds
\begin{align}
    || \pi_1-\pi_2 ||_{\cH_s}^2=\sum_{k=1}^\infty\frac{(\xi_k^{\pi_1}-\xi_k^{\pi_2})^2}{\lambda_{k}}.
\end{align}
\label{lemma:rkhs_diff}
\end{lemma}

\begin{lemma}
Let $(\xi_1^{\pi_1},\xi_1^{\pi_2}),..,(\xi_K^{\pi_1},\xi_K^{\pi_2})$ be the projection weights of $\pi_i$ onto $\mathcal{H}_s$ such that, for $p,q\in \mathbb{N}^+$, if $p>q$ then $\xi_p^{\pi_i}<\xi_q^{\pi_i}$. If there exists
a $K\in\mathbb{N}^+$ such that for all $k>K$, $|\xi_k^{\pi_1}-\xi_k^{\pi_2}| < \varepsilon^k\sqrt{\lambda_k}$ for some real $\varepsilon>0$, then the following holds
\begin{equation}
    \begin{split}
        \sum_{k=K+1}^{\infty}\frac{(\xi_k^{\pi_1}-\xi_k^{\pi_2})^2}{\lambda_k}
        &\leq \frac{\varepsilon^{2(K+1)}}{1-\varepsilon^2}.
    \end{split}
\end{equation}
\label{lemma:rkhs_geom}
\end{lemma}

The truncated embedding of size $K$ can be formed by setting the sequence of coefficients $\{\xi_k^\pi\}_{k=K}^\infty$ to zero, obtaining
\begin{equation}
    \hat{\pi}_K(t)=\sum_{k=1}^K\xi_k\omega_k(t)\;.
\end{equation}

As shown in Experiments~\ref{sec:denoising_motivation}, truncating the embedding at a given rank can have desirable properties on the returns and on their variance.

\subsection{Truncating RKHS embeddings can reduce variance of returns}
In this subsection, we show under which conditions one can expect a reduction in variance of returns of the RKHS policy, thus helping in sensitive tasks such as medical interventions or autonomous vehicles~\citep{garcia2015comprehensive}.

Any policy $\pi$ in the RKHS can be decomposed as a finite sum:
\begin{equation}
    \pi(a|s)=\underbrace{\sum_{k=1}^K\xi_k \omega_k(s,a)}_{\hat{\pi}_K}+\underbrace{\sum_{k=K}^\infty\xi_k \omega_k(s,a)}_{\varepsilon_K},\; \forall s,a\in\mathcal{S}\times \mathcal{A}.
\end{equation}

Our framework allows to derive variance guarantees by first considering the \emph{random return} variable, $Z$ of the policy $\pi$ given some state and action:
\begin{equation}
    Z(\pi|s_0,a_0)=r(S_0,A_0)+\sum_{t=1}^\infty \gamma^t r(S_t,A_t),
\end{equation}
with $S_t\sim P(\cdot|s_{t-1},a_{t-1}),A_t\sim \pi(\cdot|s_t)$ and $S_0,A_0\sim \beta$ for some initial distribution $\beta$.

Suppose we have access to some entry-wise normalization function $\sigma:\mathbb{R}\to [0,1]$. We are not concerned with the exact form of $\sigma$, but, as an example, we use the softmax function $\sigma(\vec{x}_i)=\frac{e^{\vec{x}_i}}{\sum_{j=1}^d e^{\vec{x}_j}}$.

The variance of $Z(\pi)$ over the entire trajectory is hard to compute~\citep{tamar2016learning} since it requires to solve a Bellman equation for $\mathbb{E}[Z^2(\pi)]$. Instead, we look at the variance over initial state-action pairs sampled from $\beta$. Since we do not know the distribution of $\pi(a|s)$ but only that of $(a,s)$ under $\beta$, we compute all expectations using the Law of the Unconscious Statistician~\citep[LOTUS]{ringner2009law}.

\begin{lemma}
Let $\mathbb{V}_\beta[X]=\mathbb{E}_\beta[X^2]-(\mathbb{E}_\beta[X])^2$ be the variance operator with respect to $\beta$, and let $\sigma$ be a positive-value, differentiable function.
If the following conditions are satisfied
\begin{enumerate}
    \item $\sigma'(\eta(\hat{\pi}_K))\leq \sigma'(\eta(\pi))$ (monotonically decreasing $\sigma'$ on $(0,\infty)$);
    \item $\sqrt{\frac{\mathbb{E}_\beta[\varepsilon^2_K]}{3}}\geq \mathbb{E}_\beta[\varepsilon_K]$ (second moment condition);
    \item $\mathbb{V}_\beta[\pi]\approx (\sigma'(\eta(\pi)))^2\mathbb{V}_\beta[Q^{\pi}]$ (Second order Taylor residual is small),
\end{enumerate}
then the following holds:
\begin{equation}
\begin{split}
    \mathbb{V}_\beta[Q^\pi]\geq \mathbb{V}_\beta[Q^{\hat{\pi}_K}]
\end{split}
\end{equation}
\label{lemma:variance}
\end{lemma} 

Lemma~\ref{lemma:variance} tells us the relation between the variance of returns collected by either the true or truncated policies (detailed analysis in Appendix).

In practice, Condition 1 is satisfied when $\sigma'$ is decreasing on $(0,\infty)$, rewards are strictly positive and $\eta(\hat{\pi}_K)\leq \eta(\pi)$.  
Condition 2 is not restrictive, and is, for example, satisfied by the Student's $t$ distribution with $\nu>1$ degrees of freedom. Condition 3 is a classical assumption in variance analysis~\citep{benaroya2005probability} and holds if the expansion is done near the expectation of the $Q$ function. As we show in the next section, even after the truncation step, our framework still provides performance important guarantees.

\subsection{Truncation bound}
Storing an approximation of the ground truth policy implies a trade-off between performance and number of basis functions. Our framework allows to control the difference in collected reward using the following theorem of projecting $\pi$ onto the first $K$ basis functions.

\begin{theorem}[]
Let $s\in\cS$ and $\cH_s$ be the associated RKHS. Let $\pi_1,\pi_2\in \cH_s$ be represented by $\{\xi_k^{\pi_1}\}_{k=1}^\infty$ and $\{\xi_k^{\pi_2}\}_{k=1}^\infty$ and let $\varepsilon$ such that Lemma~\ref{lemma:rkhs_geom} holds. Let $M_s>0$ be such that $|\pi_1(a|s)-\pi_2(a|s)|\leq M_s||\pi_1-\pi_2||_\cH$ for all $a\in\cA$. Let $\Delta_{\cH_s}^K=\sum_{k=1}^K\frac{ (\xi_k^{\pi_1}-\xi_k^{\pi_2})^2}{\lambda_k}$ and $\overline{\epsilon}_i = \max_{s,a} |A_{\pi_i}(s,a)|$. Then
\begin{equation*}
\begin{split}
    |\eta(\pi_2)-\eta(\pi_1)| &\leq \frac{4 |\cA|^2 \overline{\epsilon} \gamma }{(1-\gamma)^2}\\
    &\bigg\{\displaystyle{\max_{s\in\cS}}(M_s \Delta_{\cH_s}^K)^2+O(\varepsilon^{2K}\displaystyle{\max_{s\in\cS}}M_s \Delta_{\cH_s}^K)\bigg\}\\ &+\max(\overline{\epsilon}_1,\overline{\epsilon}_2)\;.
\end{split}
\end{equation*}

 \label{thm:mdp_rewards_trpo}
\end{theorem}

This theorem implies that representing any policy within the top $K$ bases of the Hilbert space yields an approximation error polynomial in $\varepsilon$ due to how the linear coefficients were picked. Since $M_s=||\kappa(\cdot,s)||_\mathcal{H}\leq \sup_{s\in\mathcal{S}}||\kappa(\cdot,s)||_\mathcal{H}<\infty$ when all function in $\mathcal{H}$ are bounded (see \cite{sun2009application}), the right hand side of Theorem~\ref{thm:mdp_rewards_trpo} is finite.

While Theorem~\ref{thm:mdp_rewards_trpo} holds in the continuous case, most projection algorithms such as Fast Fourier Transform, wavelet transform and SVD operate on a discretized version of the function. The next section describes the error made during this discretization step.

\subsection{Discretization error bound}
So far, the theoretical formulation of our algorithm was in the continuous space of states and actions. However, most efficient algorithms to project a function onto an orthonormal basis operate in the discrete space~\citep{fourier_analysis_book,daubechies1988orthonormal}. For this reason, we introduce the discretization step, which allows to leverage these highly optimized methods without sacrificing the approximation guarantees from the continuous domain.

An important step in our algorithm is the component-wise grouping of similar states. This step is crucial, since it allows us to greatly simplify the calculations of the error bounds and re-use existing discrete projection algorithms such as Fast Fourier Transform. However, when the discretization is done naively, it can result in slower computation times due to a blowup in state-action space dimensions (to mitigate this, we propose the pruning step in next section). We use an approach known as \textit{quantile bining}~\citep{naeini2015obtaining}. We assume that state components assigned to the same bin have a similar behaviour, a somewhat limiting condition which is good enough for simple environments and greatly simplifies our proposed algorithm. 

Recall that the cumulative distribution function $\Pi_X(x)=\mathbb{P}[X\leq x]$ and the corresponding quantile function $\mathcal{Q}_X(p)=\inf\{x\in \mathbb{R}:p\leq \Pi_X(x)\}$. In practice, we approximate $\mathcal{Q},\Pi$ with the empirical quantile and density functions $\hat{\mathcal{Q}},\hat{\Pi}$, respectively.

\begin{theorem}
 Let $\Pi_S,\Pi_A$ be cumulative policy functions over $\mathcal{S},\mathcal{A}$ and let $\ceil{\Pi}_S,\ceil{\Pi}_A$ be their empirical estimates discretized using quantile bining over states and actions into $b_S$ and $b_A$ clusters, respectively. Let $\mathcal{Q}_S,\mathcal{Q}_A$ be the corresponding empirical quantile functions. Then, the volume of the discretization error can be written as
 \begin{equation}
 \begin{split}
    \color{green}\delta_{\Pi,\ceil{\Pi}}\color{black}= &\sum_{i=1}^{b_S}\sum_{j=1}^{b_A}\bigg\{ \bigg|\color{orange}\int^{q^S_i}_{q^S_{i-1}}\ceil{\hat{\Pi}}_S(s)ds\color{black}-\color{purple}\frac{i}{b_S}(q^S_i-q^S_{i-1})\color{black}\bigg|\\
    & \bigg|\color{orange}\int^{q^A_{i,j}}_{q^A_{i,j-1}}\ceil{\hat{\Pi}}_{A}(a)da\color{black}-\color{purple}\frac{j}{b_A}(q^A_{i,j}-q^A_{i,j-1})\color{black}\bigg|\bigg\},
 \end{split}
\end{equation}
where $q^S_i=\ceil{\hat{\mathcal{Q}}_S}\big(\frac{i}{b_S}\big)$ and $q^A_{i,j}=\ceil{\hat{\mathcal{Q}}_{A|i}}\big(\frac{j}{b_A}\big)$.
\label{thm:discretization_thm}
\end{theorem}

Note that $\delta_{\Pi,\ceil{\Pi}}$ is similar to $||\Pi,\ceil{\Pi} ||_1$, with the only difference that the volume of each error rectangle is considered. Therefore, an argument identical to Theorem~\ref{thm:mdp_rewards_trpo} can be used to map Theorem~\ref{thm:discretization_thm} into the error space of returns.

\begin{figure}[h!]
    \centering
    \includegraphics[width=.9\linewidth]{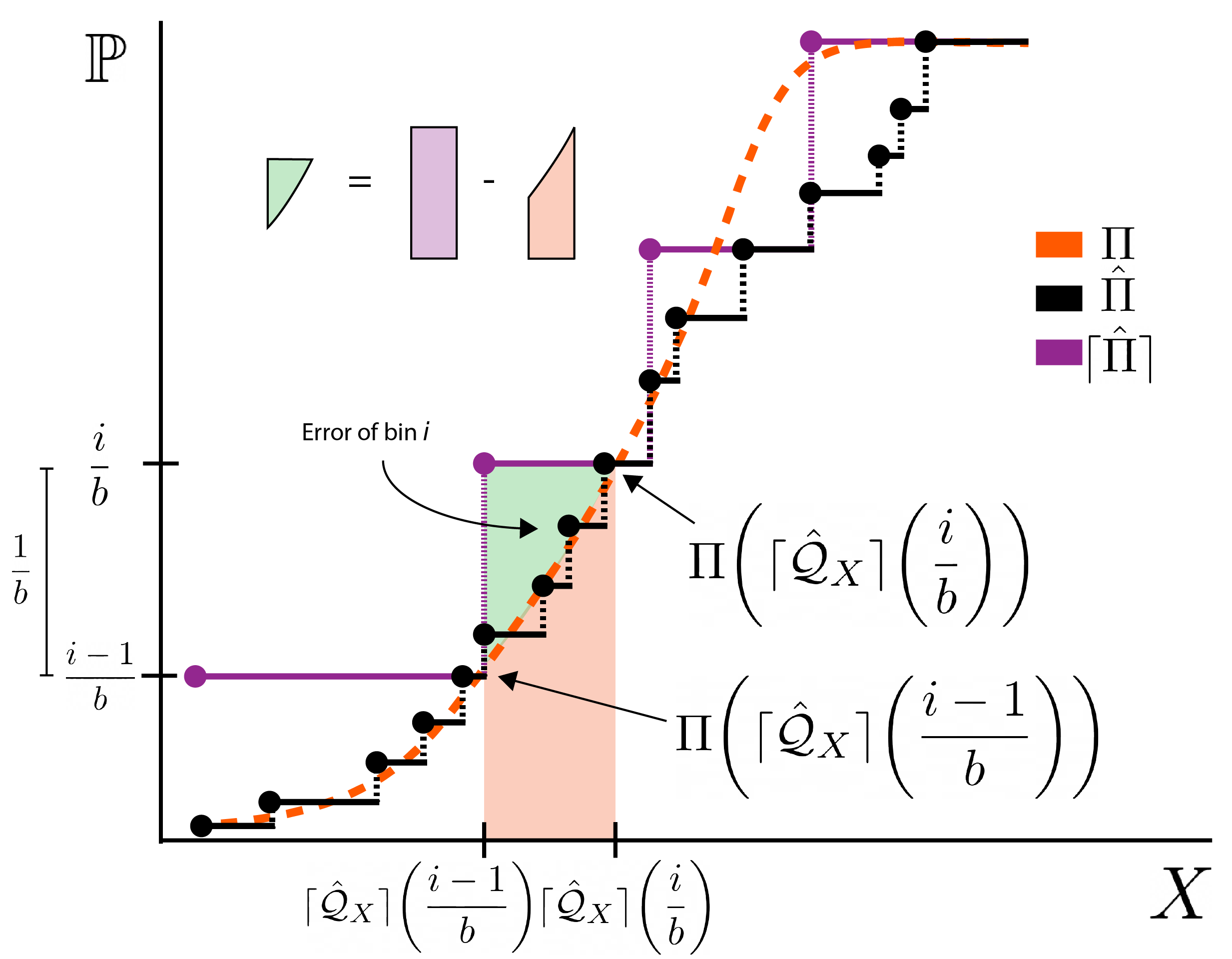}
    \caption{Schematic view of the discretization error}
    \label{fig:discretization_cartoon}
\end{figure}

Theorem~\ref{thm:discretization_thm} and Figure~\ref{fig:discretization_cartoon} show how the discretization error induced by Algorithm~\ref{algo:policy_discretization} can be computed across all states and actions in the training set. Note that samples used to compute both distribution functions are i.i.d. inter-, but not intra-trajectories. This means that the classical result from~\cite{dvoretzky1956asymptotic} might not hold.

The discretization step enables us to use powerful discrete projection algorithms, but as a side effect can drastically expand the state space. However, we observed empirically that most policies tend to cover a small subset of the state space. Using this information, we introduce a state-space pruning method based on the steady state distribution, which is addressed in the followed section.

\subsection{Pruning error bound}
The discretization step from the previous section greatly simplifies the calculations of the projection weights. However, it can potentially make the new state-action space quite large.
In this section, we introduce the pruning step based on the long-run distribution of the policy, and then quantify the impact of removing unvisited states both on the performance and coverage of the policy.
\subsubsection{Pruning unvisited states}
Since policies tend to concentrate around the optimum as training progresses~\citep{ahmed2018understanding}, pruning those states would not significantly hinder the overall performance of the embedded policy. Let $\rho_{\pi}(s)$ denote the number of times $\pi$ visits state $s$ in the long run and $N(s,a)$ the number of times action $a$ is taken from $s$ over $N$ trajectories.

This leads us to define $\pi_{pruned}$ as a policy which acts randomly under some state visitation threshold. Precisely, if $S_0=\{s:\rho_\pi(s)=0\}$ and $\mathbbm{1}(\cdot)$ is the indicator function, then
\begin{equation}
    \pi_\text{pruned}(a|s)=\{1-\mathbbm{1}_{S_0}(s)\}\frac{N(s,a)}{\rho_\pi(s)} + \mathbbm{1}_{S_0}(s)\frac{1}{b_A}\;.
    \label{eq:pi_pruned}
  \end{equation}

Pruning rarely visited states can drastically reduce the number of parameters, while maintaining high probability performance guarantees for $\pi_\text{pruned}$. The following theorem quantifies the effect of pruning states on the performances of the policy.
\begin{theorem}[Policy pruning,~\cite{spibb_estimated_baseline}]
Let $N$ be the number of rollout trajectories of the policy $\pi$, $r_{M}$ be the largest reward. With probability $1-2\delta$, the following holds: 
\begin{equation}
    |\eta(\pi)-\eta(\pi_\text{pruned})| \leq \frac{2 r_{M}}{1-\gamma}\sqrt{\frac{3|\mathcal{S}||\mathcal{A}|+4\log\frac{1}{\delta}}{2N}}\;.
\end{equation}
\label{thm:pruning}
\end{theorem}
This theorem allows one to discretize only the subset of frequently visited states while still ensuring a strong performance guarantee with high probability. 

Instead of pruning based on samples from $\rho_{\ceil{\pi}}$, which induces a computational overhead since we need to perform an additional set of rollouts, we instead prune states based on $\rho_\pi$. The approximation error induced by this switch can be understood through the scope of Thm.~\ref{thm:rho_minus_rho_tilde}.

\subsubsection{Impact of pruning on state visitation}
Any changes done to the ground truth policy such as embedding it into an RKHS will affect its stationary distribution and hence the long-run coverage of the policy. In this subsection, we provide a result on the error made on the stationary distributions as a function of error made in the original policies.

Under a fixed policy $\pi$, an MDP induces a Markov chain defined by the \textit{expected transition model} $(\mat{P}_\pi)_{ss'}=\sum_{a\in \cA} \pi(a|s)P(s'|s,a)$. In tensor form, it can be represented as $\mat{A}(\mat{T}_{(2)}\mat{\Pi})$, where $\mat{\Pi}_{as}=\pi(a|s)$ is the policy matrix, $\mat{A}=\mat{I}\odot \mat{I}$ is the Khatri-Rao product and $\mat{T}_{(2)}$ is the second matricization of the transition tensor $\mat{T}_{sas'}=P(s'|=s,a)$. 

If the Markov chain defined above is irreducible and homogeneous, then its \textit{stationary distribution} corresponds to the state occupancy probability\footnote{The existence and uniqueness of $\rho$ follow from the Perron-Frobenius theorem.} given by the principal left eigenvector of $\mat{P}_\pi$.

 The following theorem bridges the error made on the reconstruction of long-run distribution as a function of the system's transition dynamics and distance between policies.

\begin{theorem}[Approximate policy coverage]
 Let $||\cdot ||_{S_p}$ be the Schatten $p$-norm and $||\cdot||_p$ be the vector $p$-norm. If ($\mat{P}_\pi$,$|\cS|^{-1}\mat{1}_{|\cS|}$ ) and ($\mat{P}_{\pi_{pruned}}$,$|\cS|^{-1}\mat{1}_{|\cS|}$ ) are irreducible, homogeneous Markov chains, then the following holds:
 \begin{equation}
    ||\rho_\pi^\top-\rho_{\pi_{pruned}}^\top||_{S_1} \leq ||\mat{Z}||_{S_\infty} ||\mat{\Pi}-\mat{\Pi}_{pruned}||_{S_1}||\mat{T}_{(3)}||_{S_2},
\label{eq:convergence_rate}
\end{equation}
where $\mat{Z}=(\mat{I}-\mat{P}_\pi+\mat{1}_{|\cS|}\rho_\pi^\top)^{-1}$ and $\mat{1}_{|\cS|}$ is a vector of all ones.
\label{thm:rho_minus_rho_tilde}
\end{theorem}
A detailed proof can be found in the Appendix. Note that the upper bound depends on both the environment's structure as well as on the policy reconstruction quality. It is thus expected that, for MDPs with particularly large singular values of $\mat{T}_{(2)}$, the bound converges less quickly than for those with smaller singular values. See Appendix for visualizations of the bound on empirical domains.

Equiped with these tools, we are ready to state the general policy embedding bound in the RKHS setting.

\subsection{General policy embedding bound}
We are finally ready to use the previous performance bounds to state the general performance result. The difference in performance of ground-truth policy and truncated embedded policy can be decomposed as a discretization error. a pruning error and a projection error:
\begin{corollary}
Given ground truth policy $\pi$, discretized policy $\ceil{\pi}$, pruned policy $\pi_{pruned}$ and embedded policy $\hat{\pi}$, the total policy embedding bound is
 \begin{equation*}
 \begin{split}
 \bigg|\eta(\pi)-\eta(\hat{\pi}) &\bigg| \leq  \underbrace{\bigg|\eta(\pi)-\eta(\ceil{\pi}) \bigg|}_{\text{Theorem~\ref{thm:discretization_thm}}} +\underbrace{\bigg|\eta(\ceil{\pi})-\eta(\pi_{pruned})\bigg|}_{\text{Theorem ~\ref{thm:pruning}}} \\
&+\underbrace{\bigg|\eta(\pi_{pruned})-\eta(\hat{\pi})\bigg|}_{\text{Theorem~\ref{thm:mdp_rewards_trpo}}}\; .
 \end{split}
\label{thm:reconstruction_bounds_3_terms}
\end{equation*}
\end{corollary}

Tighter bounds of Thm.~\ref{thm:mdp_rewards_trpo} can be found in the literature~\cite{giardina1972bounds,rakhlin2005risk} for specific basis functions. 

In the next section, we propose a practical algorithm which integrates all three steps.

%% file: algorithm.tex
We first discuss the construction of $\ceil{\pi}$ via quantile discretization. The idea consists in grouping together similar state-action pairs (in term of visitation frequency). To this end, we use the quantile state visitation function $\mathcal{Q}_S$ and the state visitation distribution function $\Pi_S$, as well as their empirical counterparts, $\hat{\mathcal{Q}}_{S}$ and $\hat{\Pi}_S$. Quantile and distribution functions for the action space are defined analogously.
 
\begin{algorithm}
\SetAlgoLined
\KwIn{Policy $\pi$, number of state bins $b_S$, number of action bins $b_A$}
\KwResult{Discrete policy $\ceil{\pi}$}
 
 Collect a set of states $S\subseteq  \cS$ and set of actions $A \subseteq \cA$ via rollouts of $\pi$\;
 
 Build the empirical c.d.f. $\hat{\Pi}_{s}$ from set $S$\;
 
 Build the empirical c.d.f. $\hat{\Pi}_{a}$ from set $A$\;
 
 Find numbers $i_1,..,i_{b_S}$ s.t. $\hat{\Pi}_s(i_l)-\hat{\Pi}_s(i_{l-1})=\frac{1}{b_S}$ using $\hat{\mathcal{Q}}_s$ for all $s\in S,l=1,..,b_S$\;
 
 Find numbers $j_1,..,j_{b_A}$ s.t. $\hat{\Pi}_a(j_l)-\hat{\Pi}_a(j_{l-1})=\frac{1}{b_A}$ using $\hat{\mathcal{Q}}_a$ for all $a\in S,l=1,..,b_A$\;
 
 \If{$i_{l-1}\leq s \leq i_{l}$ and $j_{l'-1}\leq a \leq j_{l'}$}{
    Assign $(s,a)$ to $(l,l')$\;
 }
 Set $\ceil{\pi}_{ll'}$ to $\pi(\frac{i_{l'-1}+i_{l'}}{2},\frac{i_{l-1}+i_{l}}{2})$
 \caption{Quantile discretization}
 \label{algo:policy_discretization}
\end{algorithm}

Algorithm~\ref{algo:policy_discretization} outlines the proposed discretization process allowing one to approximate any continuous policy~(e.g., computed by a neural network) by a 2-dimensional table indexed by discrete states and actions. We use quantile discretization in order to have maximal resolution in frequently visited areas. It also allows for slightly faster sampling during rollouts, since the probability of falling in each bin is uniform.

\begin{algorithm}
\SetAlgoLined
\KwIn{Policy $\pi$, number of components $K$, basis $\omega=\{\omega_k\}_{k=1}^\infty$}
\KwResult{A set of coefficients $\xi_1,...,\xi_K$}
 1. Rollout $\pi$ in the environment to estimate $\hat{\mathcal{Q}}_s,\hat{\mathcal{Q}}_a$ (empirical quantile functions)\;
 
 2. Project $\pi$ onto lattice to obtain $\ceil{\pi}$ using Alg.~\ref{algo:policy_discretization}\;
 
 3. Prune $\ceil{\pi}$ with $\hat{\mathcal{Q}}_s,\hat{\mathcal{Q}}_a$ using Eq.~\ref{eq:pi_pruned} \;
 
 4. Project $\ceil{\pi}$ onto first $K$ elements of $\omega$ using one of the projection algorithms described in Section 2.2\;
 
 \caption{RKHS policy embedding}
 \label{algo:joint_reconstruction}
\end{algorithm}

Since in practice working with continuous RKHS projections is cumbersome, we use Algorithm~\ref{algo:policy_discretization} in order to project the continuous policy $\pi$ onto a discrete lattice, which is then projected onto the Hilbert space\footnote{Python pseudocode can be found in Appendix~\ref{sec:python_code}}. A natural consequence of working with embedded policies means that "taking actions according to $\hat{\pi}_K$" reduces to importance sampling of actions conditional on state bins (discussed in Appendix~\ref{sec:acting_tabular}).

%% file: experiments.tex
In this section, we consider a range of control tasks: bandit turntable, Pendulum and Continuous Mountain Car (CMC) from OpenAI Gym \citep{OpenAIgym}. All experimental details can be found in Appendix~\ref{app:exp_parameters}.

In Pendulum and CMC, we omit GMM and fixed basis GMM methods, since the expectation-maximization algorithm runs into stability problems when dealing with a high number of components. Fixed-basis GMM serves as a baseline to benchmark optimization issues in the Expectation-Maximization (EM) algorithm for GMMs.  Moreover, we use SVD for low-rank matrix factorization and the fourth order Daubechies wavelet~\citep{daubechies1988orthonormal} (DB4) as an orthonormal basis alternative to Fourier (DFT).

\subsection{Bandit turntable}

\begin{figure}[h!]
    \centering
    \includegraphics[width=\linewidth]{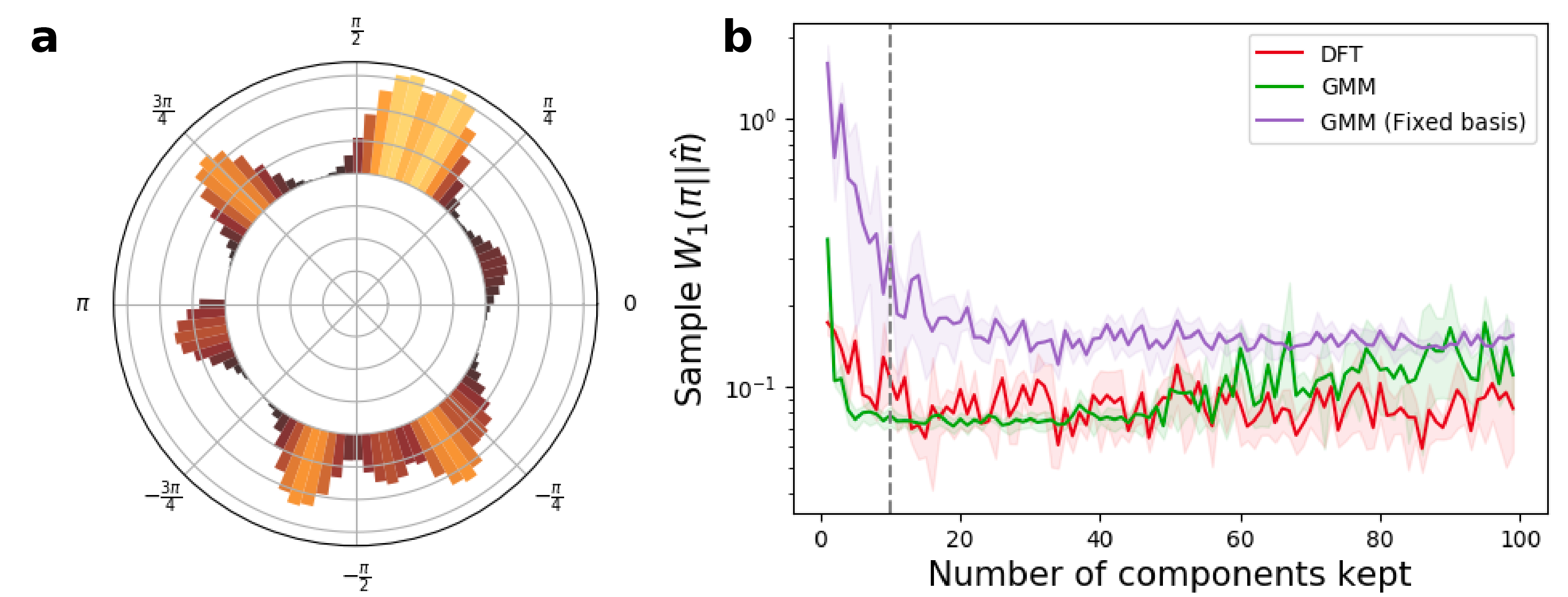}
    \caption{Bandit turntable environment in \textbf{(a)} polar coordinates. Magnitude of modes is proportional to the reward (lighter is higher). Actions are absolute angles in interval $ [-\pi,\pi]$. \textbf{(b)} shows the $W_1$ distance between ground truth and order $K$ approximations. Dotted line indicates the number of modes in the ground truth density. }
    \label{fig:bandit_turntable}
\end{figure}

We evaluate our framework in the multi-armed bandit~\cite[MAB]{slivkins2019introduction} inspired from \cite{fourier_pg} (see Figure~\ref{fig:bandit_turntable}a). Results are reported in Figure~\ref{fig:bandit_turntable}b via the average Wasserstein-1 metric, defined as $\mathbb{E}_s[W_1(\pi(\cdot|s),\hat{\pi}(\cdot|s))]$. From the figure, we can observe that GMM has more trouble converging when we keep more than 60 components. This is due to the EM's stability issue when dealing with large number of components. While fixed basis GMM struggles to accurately approximate the policy, DFT shows a stable performance after a threshold of 10 components. The appendix contains additional results in the MAB setting motivated by recent advances~\citep{dudik2011efficient,foster2018practical}, and a simplified truncation lemma for average rewards.

\subsection{Continuous Mountain Car} 
\begin{figure}[h!]
    \centering
    \includegraphics[width=0.7\linewidth]{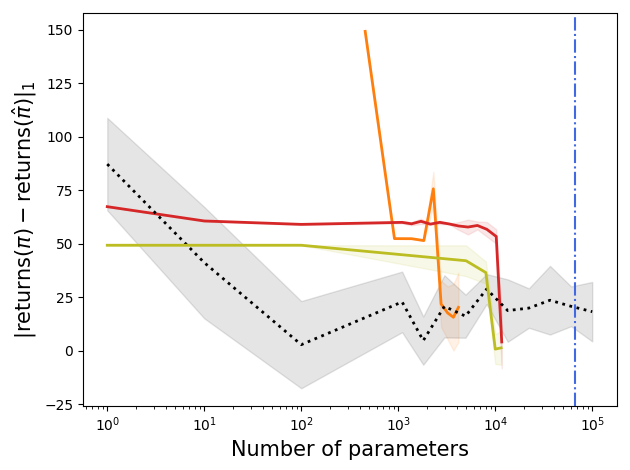}
    \caption{Difference in returns between embedded and ground truth policies for the Mountain Car task with respect of the number of parameters used.}
    \label{fig:difference_return_mountain_car}
\end{figure}

We compare all embedding methods with a maximum likelihood estimate (MLE) of $\pi$, which consists of approximating $\mathbb{E}[\pi_K(\cdot|s)]\approx \frac{1}{K}\sum_{k=1}^K A_k,\; A_k\sim\pi(\cdot|s)$. The MLE rate of convergence to $\pi$ is $O(\sqrt{K})$, which grounds the convergence rate of other methods. As shown in Figure~\ref{fig:difference_return_mountain_car} over 10 runs, DFT, SVD and DB4 reach same returns as the MLE method. Note that DB4 shows slightly better performance than DFT. 
 
Due to space constraints, Appendix Figure~\ref{fig:cmc_large_fig} contains insights on the pruning and discretization errors.

\subsection{Pendulum}
\begin{figure}[h!]
    \centering
    \includegraphics[width=\linewidth]{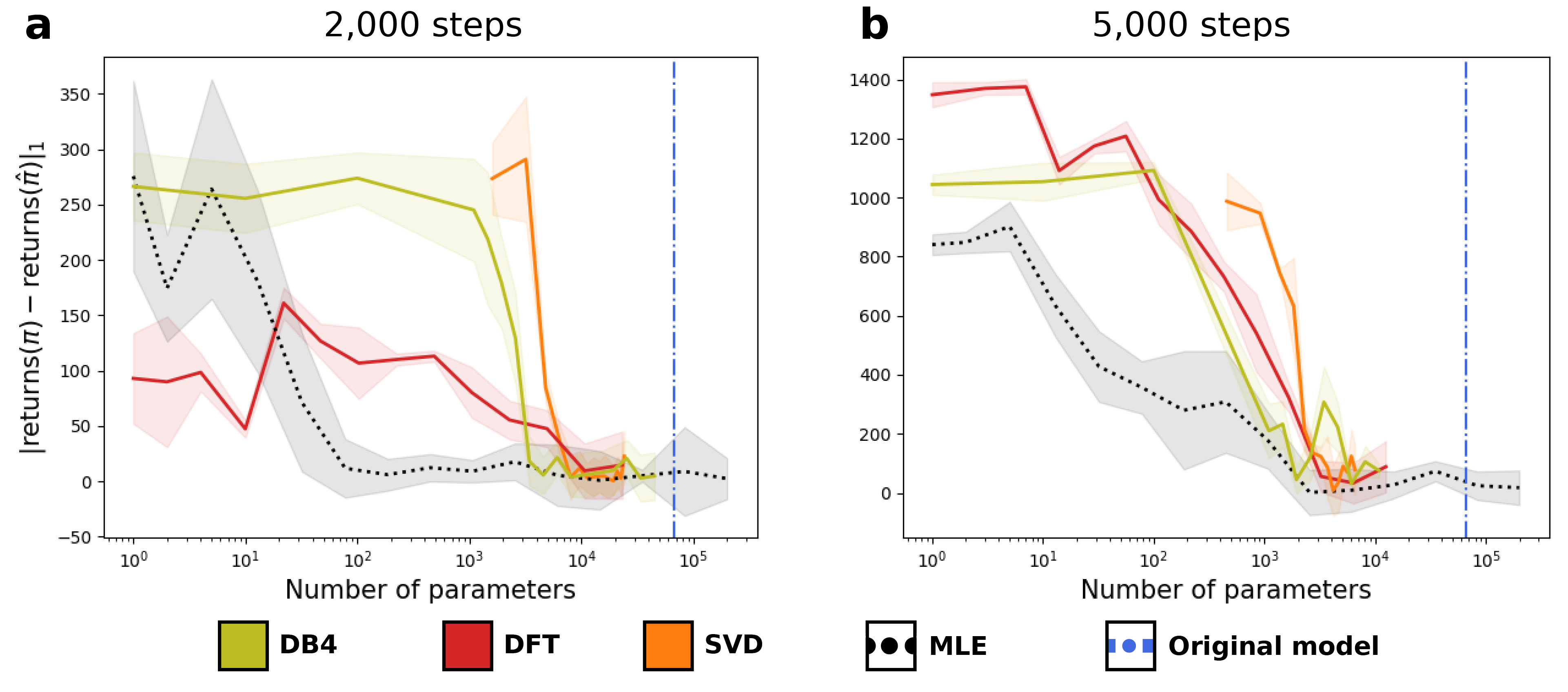}
    \caption{Absolute difference in returns collected by discretized and reconstructed Gaussian policies for \textbf{(a)} 2,000 steps and \textbf{(b)} 5,000 steps in the Pendulum task, averaged over 10 trials. Blue dots represent the number of parameters of the neural network policy. SVD, DFT and DB4 projections need an order of magnitude less in term of parameters to reconstruct the original policy.}
    \label{fig:rewards_pendulum}
\end{figure}

The same experimental setup as for Mountain Car applies to this environment.
 As shown in Figure~\ref{fig:rewards_pendulum} over 10 runs, when the true policy has converged to a degenerate (optimal) distribution ($5,000$ steps), all methods show comparable performance (in terms of convergence). DFT shows better convergence at early stages of training ($2,000$ steps), that is when the true policy has a large variance. Refer to Appendix Figure~\ref{fig:cartoon_pendulum}-\ref{fig:radial_boltzmann_pendulum} for visualizations of the true policies.
  
\begin{table}[h!]
    \centering
    \begin{tabular}{c|cc}
    \toprule
        Task & $\sqrt{\mathbb{V}[\eta(\pi)]}$ & $\sqrt{\mathbb{V}[\eta(\hat{\pi}_K)]}$ \\
        \midrule
        Pendulum (2k) & 114.51 & 84.75\\
        Pendulum (3k) & 333.06 & 191.83\\
        Pendulum (4k) & 363.6 & 273.1\\
        Pendulum (5k) & 182.2 & 163.3\\
        CMC & 28.1 & 11.0\\
        \bottomrule
    \end{tabular}
    \caption{Variance of the collected returns in Pendulum-v0 under various policies over 100 trials, with $K$ set to half of its maximum value (i.e. $K=b_Sb_A/2$).}
    \label{tab:variance}
\end{table}

Table~\ref{tab:variance} shows the standard deviation in returns collected by true and truncated policies over 100 trials. The truncated policies systematically has standard deviation at least as large as the one of the true policy.

%% file: conclusion.tex
In this work, we introduced a general framework for representing policies for reinforcement learning tasks. It allows to represent any continuous policy density by a projection onto the $K$ first basis functions of a reproducing kernel Hilbert space. We showed theoretically that the performance of this embedded policy depends on the discretization, pruning and projection algorithms, and we provide upper bounds for these errors. In addition, by projecting to a lower dimensional space, our framework provides a more stable and robust reconstructed policy. Finally, we conducted experiments to demonstrate the behaviour of SVD, DFT, DB4 and GMM basis functions on a set of classical control tasks, supporting our performance guarantees. Moreover, our experiments show a reduction in the variance of the returns of the reconstructed policies, resulting a more robust policy. A natural extension to our framework would be to directly operate in the continuous space, avoiding the discretization step procedure.

%% file: appendix.tex
\subsection*{Reproducibility Checklist}

We follow the reproducibility checklist~\textit{"The machine learning reproducibility checklist"} and point to relevant sections explaining them here.
\\
For all algorithms presented, check if you include:\\
\begin{itemize}
    \item \textbf{A clear description of the algorithm, see main paper and included codebase.}
    The proposed approach is completely described by Alg.~\ref{algo:joint_reconstruction}.
\item \textbf{An analysis of the complexity (time, space, sample size) of the algorithm.}
The space complexity of our algorithm depends on the number of desired basis. It is $4K$ for a 1-dimension $K-$component GMM with diagonal covariance, $mK+K^2+nK$ for a $K-$truncated SVD of an $m\times n$ matrix, and only $K$ for the wavelet and Fourier bases. Note that a simple implementation of SVD requires to find the three matrices $\mat{U},\mat{D},\mat{V}^\top$ before truncation.

The time complexity depends on the reconstruction, pruning and discretization algorithms, which involves conducting rollouts w.r.t. policy $\pi$, training a quantile encoder on the state space trajectories, and finally embedding the the pruned matrix using the desired algorithm. We found that the Gaussian mixture is the slowest to train, due to shortcomings of the expectation-maximization algorithm.

\item \textbf{A link to a downloadable source code, including all dependencies.} The code is included with Supplemental Files as a zip file; all dependencies can be installed using Python's package manager. Upon publication, the code would be available on Github. Additionally, we include the model's weights as well as the discretized policy for \texttt{Pendulum-v0} environment.
\end{itemize}
For all figures and tables that present empirical results, check if you include:
\begin{itemize}
    \item \textbf{A complete description of the data collection process, including sample size.} We use standard benchmarks provided in OpenAI Gym (Brockman et al., 2016).
    \item \textbf{A link to downloadable version of the dataset or simulation environment.} Not applicable.
\item \textbf{An explanation of how samples were allocated for training / validation / testing.} We do not use a training-validation-test split, but instead report the mean performance (and one standard deviation) of the policy at evaluation time across 10 trials.
\item \textbf{An explanation of any data that were excluded.} 
The only data exclusion was done during policy pruning, as outlined in the main paper.
\item \textbf{The exact number of evaluation runs.} 10 trials to obtain all figures, and 200 rollouts to determine $\rho_\pi$.
\item \textbf{A description of how experiments were run.} See Section Experimental Results in the main paper and didactic example details in Appendix.
\item \textbf{A clear definition of the specific measure or statistics used to report results.} Undiscounted returns across the whole episode are reported, and in turn averaged across 10 seeds. Confidence intervals shown in Fig.~\ref{fig:difference_return_mountain_car} were obtained using the pooled variance formula from a difference of means $t-$test.
\item \textbf{Clearly defined error bars.} Confidence intervals are always mean$\pm\;1$ standard deviation over 10 trials.
\item \textbf{ A description of results with central tendency (e.g. mean) and variation (e.g. stddev)}. All results use the mean and standard deviation.
\item \textbf{ A description of the computing infrastructure used.} All runs used 1 CPU for all experiments with $8$Gb of memory. 
\end{itemize}

\newpage

\subsection{Bandit example}
 An $N$-armed bandit has a reward distribution such that $r(i)=\frac{1}{\sqrt{2\pi}}\exp(-(i-N/2)^2/2)+U_i$ for $i=1,..,N$, and stationary i.i.d. measurement noise $U$, with the additional restriction that $\mathbb{V}(U)=\lambda^2$, i.e. the signal-to-noise ratio (SNR) simplifies to $\text{SNR}=\frac{N}{2\lambda}$.
 
 Running any suitable policy optimization method~\cite{dudik2011efficient,foster2018practical} then produces a policy $\pi$ which regresses on both the signal and the noise (blue curve in Figure~\ref{fig:motivating_example}). However, if the class of regressor functions is overparameterized, then the policy will also capture the noise, which in turn will lead to suboptimal collected rewards. If regularization is not an option (e.g. the training data was deleted due to privacy requirements as discussed in~\cite{sen2014bootstrapping}), then separating signal from noise becomes more challenging.

Our approach involves projecting the policy into a Hilbert space in which noise is captured via fine-grained basis functions and therefore can be eliminated via truncation of the basis. Figure~\ref{fig:motivating_example} illustrates the hypothetical bandit scenario, and the behavior of our approach. Executing the policy with 2 components will yield a larger reward signal than executing the original policy, because the measurement noise is removed via truncation.

\subsection{Convergence results for Fourier approximation and GMM mixture} \label{UDA_rate
}
\paragraph{Rate of convergence of Fourier approximation}
Let $\pi_K^{\text{Fourier}}$ denote the density approximated by the first $K^{\text{th}}$ Fourier partial sums~\cite{fourier_analysis_book}, then a result from~\cite{jackson_1930} shows that
\begin{equation}
    ||\pi-{\pi}_K^\text{Fourier}||_1=O(K^{-1})\;.
    \label{eq:jackson_l1_dft}
\end{equation}
More recent results~\cite{giardina1972bounds} provide even tighter bounds for continuous and periodic functions with $m-1$ continuous derivatives and bounded $m^\text{th}$ derivative. In such case,
\begin{equation}
    ||\pi-\pi_K^\text{Fourier}||_{\infty}=O(K^{-(2m+1)})\;.
    \label{eq:giardina_linf_dft}
\end{equation}
\paragraph{Rate of convergence of mixture approximation}
Let $\pi_K^{\text{MM}}$ denote a (finite) $K-$component mixture model. A result from~\cite{rakhlin2005risk} shows the following result for $\pi_K^\text{MM}$ in the class of mixtures with $K$ marginally-independent scaled density functions and $\pi$ in the class of lower-bounded probability density functions:
\begin{equation}
    KL(\pi||\pi_K^\text{MM})=O(K^{-1}+n^{-1}),
    \label{eq:rakhlin_kl_mm}
\end{equation}
where $n$ is the number of i.i.d. samples used in the learning of $\pi_K^\text{MM}$.

The constants hidden inside the big-O notation depend on the nature of the function and can become quite large, which can explain differences in empirical evaluation.

\subsection{Discretization of continuous policies}
\label{sec:discretization}
Consider the case when $f$ is a continuous, positive and decreasing function with a discrete sequence $a_n=f(n)$. For example, the reconstruction error $W_1(\Pi,\hat{\Pi})$ as well as difference in returns $|\eta(\pi)-\eta(\hat{\pi})|$ fall under this family. Then, $\int_{0}^\infty f(t)dt<\infty$ implies that $\sum_{i=0}^\infty a_n<\infty$. Under these assumptions, convergence guarantees (e.g. on monotonically decreasing reconstruction error) in continuous space imply convergence in the discrete (empirical) setting. Hence, we operate on discrete spaces rather than continuous ones.

All further computations of distance between two discretized policies will have to be computed over the corresponding discrete grid $(i,j), i=1,..,b_S\;,\;j=1,..,b_A$. One can look at the average action taken by two MDP policies at state $s$:
\begin{equation}
\begin{split}
    \bigg|\mathbb{E}_{a \sim \pi}[a|s]-\mathbb{E}_{a \sim \tilde{\pi}}[a|s]\bigg|&\leq W_1(\Pi_s,\tilde{\Pi}_s)
    \label{eq:w1_discretized_mdp}
\end{split}
\end{equation}
This relationship is one of the motivations behind using $W_1$ to assess goodness-of-fit in the experiments section.

Naively discretizing some function $f$ over a fixed interval $[a,b]$ can be done by $n$ equally spaced bins. To pass from a continuous value $f(x)$ to discrete, one would find $i\in 1,..,n$ such that $\frac{b-a}{n}i \leq x \leq \frac{b-a}{n}(i+1)$. However, using a uniform grid for a non-uniform $f$ wastes representation power. To re-allocate bins in low density areas, we use the quantile binning method, which first computes the cumulative distribution function of $f$, called $F$. Then, it finds $n$ points such that the probability of falling in each bin is equal. Quantile binning can be see as uniform binning in probability space, and exactly corresponds to uniform discretization if $f$ is constant (uniform distribution).\\
Below, we present an example of discretizing 4,000 sample points taken from four $\mathcal{N}(\mu_i,0.3)$ distributions, using the quantile binning.

\begin{figure}[h!]
    \centering
    \includegraphics[width=0.5\linewidth]{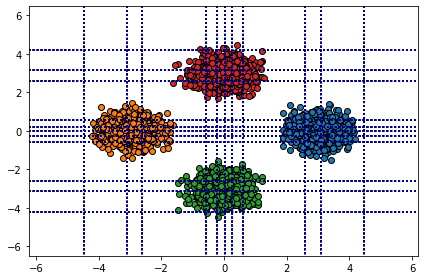}
    \caption{Quantile binning for the four Gaussians problem, using 10 bins per dimension.}
    \label{fig:quantile_binning_4_gaussians}
\end{figure}

In Figure~\ref{fig:quantile_binning_4_gaussians}, we are only allowed to allocate 10 bins per dimension. Note how the grid is denser around high-probability regions, while the furthermost bins are the widest. This uneven discretization, if applied to the policy density function, allows the agent to have higher detail around high-probability regions.\\
In Python, the function \texttt{sklearn.preprocessing.KBinsDiscretizer} can be used to easily discretize samples from the target density.\\
\begin{figure}[h!]
    \centering
    \includegraphics[width=0.5\linewidth]{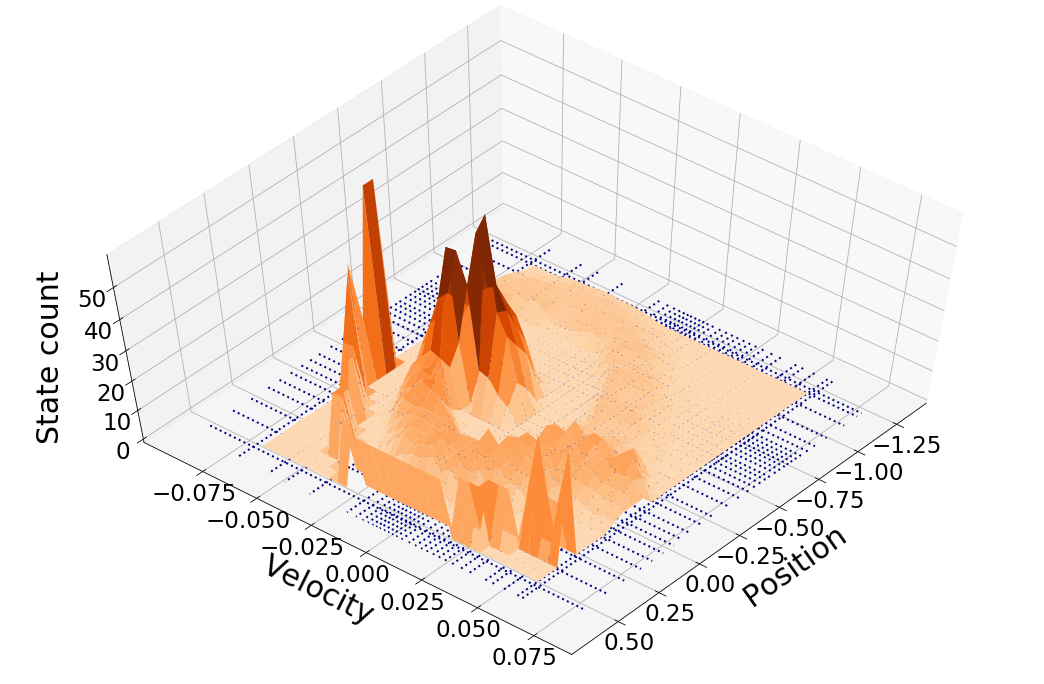}
    \caption{Quantile binning of unnormalized state visitation counts of DQN policy in the discrete Mountain Car task. Bins closer to the starting states are visited more often than those at the end of the trajectory. Blue dots represents the bins limits.}
    \label{fig:quantile_binning_MC}
\end{figure}

Since policy densities are expected to be more complex than the previous example, we analyze quantile binning in the discrete Mountain Car environment. To do so, we train a DQN policy until convergence (i.e. collected rewards at least $-110$), and perform 500 rollouts using the greedy policy. Trajectories are used to construct a 2-dimensional state visitation histogram. At the same time, all visited states are given to the quantile binning algorithm, which is used to assign observations to bins.

Figure~\ref{fig:quantile_binning_MC} presents the state visitation histogram with $n_S=30$ bins, where color represents the state visitation count. Bins are shown by dotted lines.

\subsection{Acting with tabular policies}
\label{sec:acting_tabular}
In order to execute the learned policy, one has to be able to (1) generate samples conditioned on a given state and (2) evaluate  the log-probability of state-action pairs. Acting with a tabular density can be done via various sampling techniques. For example, importance sampling of the atoms corresponding to each bin is possible when $b_A$ is small, while the inverse c.d.f. method is expected to be faster in larger dimensions. The process can be further optimized on a case-per-case basis for each method. For example, sampling directly from the real part of a characteristic function can be done with the algorithm defined in \cite{devroye1986automatic}

Furthermore, it is possible to use any of the above algorithms to jointly sample $(s,a)$ pairs under the assumption that states were discretized by quantiles. First, uniformly sample a state bin and then use any of the conditional sampling algorithms to sample action $a$. Optionally, one can add uniform noise (clipped to the range of the bin) to the sampled action. This naive trick would transform discrete actions into continuous approximations of the policy network. 

 
 
 
 

\subsection{Bandit motivating example}
\label{sec:denoising_motivation}
\begin{figure}[h!]
    \centering
    \includegraphics[width=\linewidth]{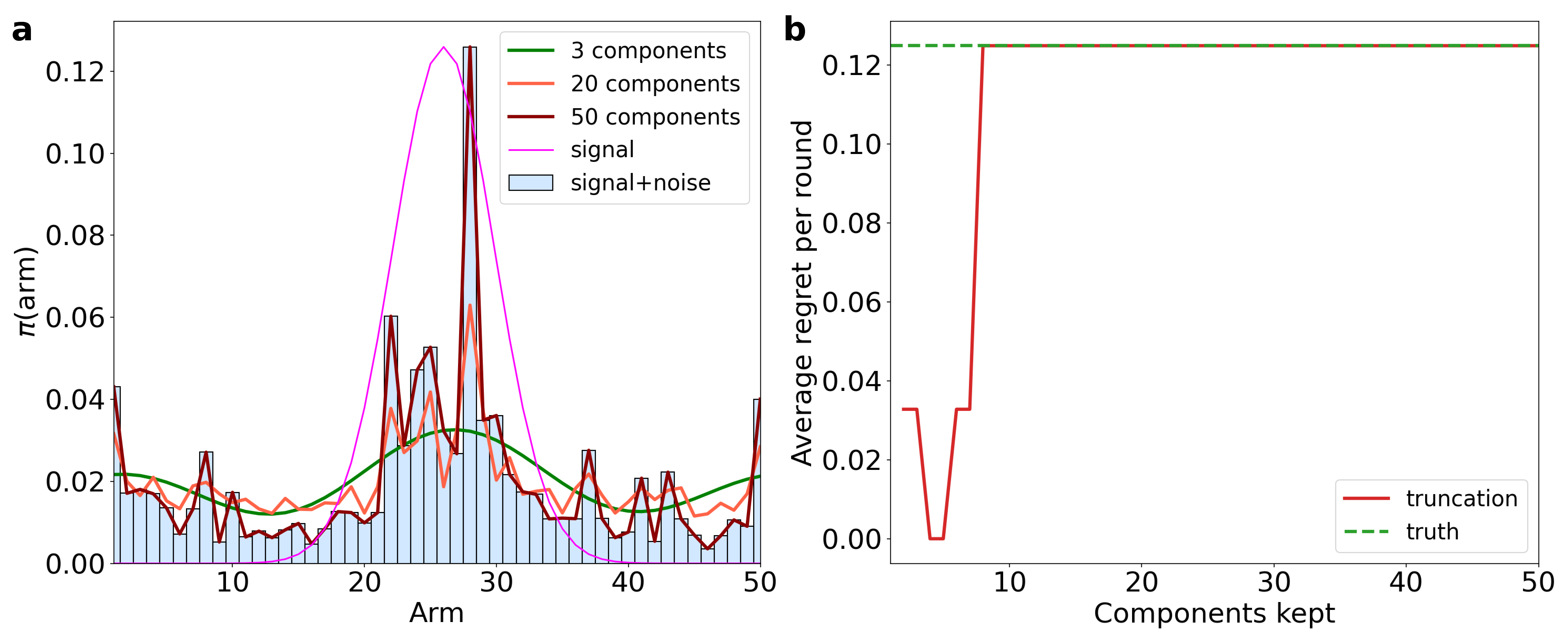}
    \caption{\textbf{(a)} Motivating bandit example with ground truth signal, noisy signal and truncated policy, and \textbf{(b)} rewards collected by greedy (i.e. the mode) truncated and true policies}
    \label{fig:motivating_example}
\end{figure}

We observe that truncating the embedded policy can have a denoising effect, i.e. boosting the signal-to-noise ratio (SNR).
As a motivation, consider a multi-armed bandit setting~\citep{slivkins2019introduction} with reward distribution  $r(i)=\frac{1}{\sqrt{2\pi}}\exp(-(i-N/2)^2/2)+U_i$ for $i=1,..,N$, and stationary measurement noise $U_i$. By running any suitable policy optimization method~\citep{dudik2011efficient,foster2018practical}, we obtain a policy $\pi$ which regresses on both the signal and the noise. Figure~\ref{fig:motivating_example} illustrates the hypothetical bandit scenario. Applying DFT followed by truncation produces $\pi_K$, some of which are shown in Figure~\ref{fig:motivating_example}a. Sampling actions from $\pi_K$ yields a lower average regret up to K=10, because the measurement noise is truncated and the correct signal mode is recovered.

The bandit performance bound can be stated as follows:
\begin{lemma}
Let $\cH$ be an RKHS, $\pi_1,\pi_2\in \cH$ be represented by $\{\xi_k^{\pi_1}\}_{k=1}^\infty$ and $\{\xi_k^{\pi_2}\}_{k=1}^\infty$ and $r: \mathcal{A} \rightarrow \mathbb{R}$ be the bandit's reward function. Let $M>0$ be such that $|\pi_1(a)-\pi_2(a)|\leq M||\pi_1-\pi_2||_\cH$ for all $a\in\cA$ and let $p,q>0$ s.t. $\frac{1}{p}+\frac{1}{q}=1$. Then
\begin{equation*}
    |\eta(\pi_1)-\eta(\pi_2)| \leq |\cA|^{1/q}M||r||_p\sum_{k=1}^\infty\frac{(\xi_k^{\pi_1}-\xi_k^{\pi_2})^2}{\lambda_{k}}
\end{equation*}
\label{lemma:bandit_reward_approximation}
\end{lemma}
Lemma~\ref{lemma:bandit_reward_approximation} guarantees that rewards collected by $\pi_2$ will depend on the distance between $\pi_1$ and $\pi_2$. The proof can be found in the next section.

\subsection{Proofs}

\begin{proof}(Lemma~\ref{lemma:variance})
Throughout the proof, we will be using the expectation $\mathbb{E}_\beta$ which can be seen as the classical inner product weighted by the initial state-action distribution
$$\langle f,g\rangle_\beta=\int_{s,a}f(s,a)g(s,a)\beta(s,a)d(sa)$$ over the domain of $s,a$.

 If we take a uniform initialization density, i.e. $\beta(s,a)\propto(|\mathcal{S}||\mathcal{A}|)^{-1}$, then the inner product simplifies to 
 $$
     \langle f,g \rangle_\beta = (|\mathcal{S}||\mathcal{A}|)^{-1} \langle f,g\rangle,
$$
which will be used further on to simplify the covariance term between orthonormal basis functions.

After a first-order Talor expansion of variances around $\eta(\pi)$ and $\eta(\hat{\pi}_K)$ described in~\cite{benaroya2005probability}, we obtain
\begin{equation}
\begin{split}
\mathbb{V}_\beta[\pi(a|s)]
&=\{\sigma'(\eta(\pi))\}^2\mathbb{V}_\beta[Q^\pi(s,a)]+E_2^\pi\\
\mathbb{V}_\beta[\hat{\pi}_K(a|s)]
&=\{\sigma'(\eta(\hat{\pi}_K))\}^2\mathbb{V}_\beta[Q^{\hat{\pi}_K}(s,a)]+E_2^{\hat{\pi}_K},
\end{split}    
\end{equation}
where $E^\pi_2$ is the Taylor residual of order 2 for policy $\pi$. This identity is very widely used in statistics, especially for approximating the variance of functions of statistics through the \emph{delta method}~\citep{wolter2007introduction}.

Expanding the variance of sum of correlated random variables, we obtain
\begin{equation}
    \begin{split}
        \mathbb{V}_\beta[\pi(a|s)]&=\mathbb{V}_\beta[\hat{\pi}_K(a|s)]+\mathbb{V}_\beta[\varepsilon_K(a|s)]
        +2\mathbb{V}_\beta[\hat{\pi}_K(a|s),\varepsilon_K(a|s)]
    \end{split}
\end{equation}

The covariance term $\mathbb{V}_\beta[\hat{\pi}_K(a|s),\varepsilon_K(a|s)]$ can be decomposed as $\mathbb{E}_\beta[\hat{\pi}_K(a|s)\varepsilon_K(a|s)]-\mathbb{E}_\beta[\hat{\pi}_K(a|s)]\mathbb{E}_\beta[\varepsilon_K(a|s)]$.
 
 Then, the second moment expression of the covariance becomes zero with respect to this inner product, due to the use of orthonormal basis functions. We use the dominated convergence theorem which lets the order of summation be changed for $\sum_{j=K}^\infty \xi_j\omega_j(s,a)$, which greatly simplifies the calculations.

\begin{equation}
\begin{split}
    \mathbb{E}_\beta[\hat{\pi}_K(a|s)\varepsilon_K(a|s)]&=\mathbb{E}_\beta[\sum_{k=1}^K\xi_k\omega_k(s,a)\sum_{j=K}^\infty\xi_j\omega_j(s,a)]\\
    &=\sum_{k=1}^K\sum_{j=K}^\infty\xi_k\xi_j\mathbb{E}_\beta[\omega_k(s,a)\omega_j(s,a)] \\
    &=\sum_{k=1}^K\sum_{j=K}^\infty\xi_k\xi_j\langle\omega_k,\omega_j\rangle_\beta\\
    &=0 \quad  (\omega_k(s,a) \text{ and } \omega_j(s,a) \text{ are orthogonal })
\end{split}
\end{equation}

Combining both expressions yields the relationship between the variance of the truncated policy and that of the expected returns.

\begin{equation}
\begin{split}
    \mathbb{V}_\beta[Q^\pi]&=
    \bigg[\frac{\{\sigma'(\eta(\hat{\pi}_K))\}^2\mathbb{V}_\beta[Q^{\hat{\pi}_K}]- 2\mathbb{E}_\beta[\hat{\pi}_K]\mathbb{E}_\beta[\varepsilon_K]+\mathbb{V}_\beta[\varepsilon_K]+E_2^{\hat{\pi}_K} }{\{\sigma'(\eta(\pi))\}^2}\bigg]+E_2^\pi\\
    &=\bigg[\frac{\{\sigma'(\eta(\hat{\pi}_K))\}^2\mathbb{V}_\beta[Q^{\hat{\pi}_K}]- 2\mathbb{E}_\beta[\hat{\pi}_K]\mathbb{E}_\beta[\varepsilon_K]+\mathbb{V}_\beta[\varepsilon_K] }{\{\sigma'(\eta(\pi))\}^2}\bigg]+E_2^\pi+\frac{E_2^{\hat{\pi}_K}}{\{\sigma'(\eta(\pi))\}^2}\\
\end{split}
\end{equation}

If the Taylor expansion is performed in a neighborhood of $\eta(\pi)$ and $\eta(\hat{\pi}_K)$, we can then consider the error terms as sufficiently small and neglect them, as is often the case in classical statistics~\citep{benaroya2005probability}.

\begin{equation}
\begin{split}
    \mathbb{V}_\beta[Q^\pi]
    &\approx\bigg(\frac{\sigma'(\eta(\hat{\pi}_K)) }{\sigma'(\eta(\pi))}\bigg)^2\mathbb{V}_\beta[Q^{\hat{\pi}_K}]+\frac{ \mathbb{V}_\beta[\varepsilon_K]-2\mathbb{E}_\beta[\hat{\pi}_K]\mathbb{E}_\beta[\varepsilon_K]}{\{\sigma'(\eta(\pi))\}^2}
\end{split}
\end{equation}


For the variance of the returns of the true policy to be not lower than the variance of returns of the truncated policy, the coefficient of the first term on the right hand side must be at most 1, and the second term must be positive. Note that these are sufficient but not necessary conditions.

Precisely, the following conditions must be satisfied:
\begin{enumerate}
    \item $\sigma'(\eta(\hat{\pi}_K))\leq \sigma'(\eta(\pi))$;
    \item $\sqrt{\frac{\mathbb{E}_\beta[\varepsilon^2_K]}{3}}\geq \mathbb{E}_\beta[\varepsilon_K]$. This is an assumption on the second moment behavior of the policy and is satisfied by, for example, the Student's $t$ distribution with degrees of freedom $\nu>1$.
    \item The Taylor error terms $E^\pi_2$ and $E^{\hat{\pi}_K}_2$ are small since the expansion is performed around $\eta(\pi)$ and $\eta(\hat{\pi}_K)$, respectively.
\end{enumerate}
 That is, under various initializations, truncation of policy coefficients can be beneficial by reducing the variance.

\end{proof}

\begin{proof}(Theorem~\ref{thm:discretization_thm})
We want to quantify the difference between the continuous cdf $\Pi$ and the step-function defined by the quantile binning strategy, $\floor{\Pi}$, with $b$ bins. Notice that the error between $\Pi$ and $\floor{\Pi}$ in bin $i$ can be written as 
\begin{equation}
    \delta_i=\int^{q_i}_{q_{i-1}}\Pi(x)dx-\frac{q_i-q_{i-1}}{b},
\end{equation}
where $q_i=\hat{\mathcal{Q}}\big(\frac{i}{b}\big)$.

The total error across all bins is therefore
\begin{equation}
    \delta=\sum_{i=1}^b \bigg[\int^{q_i}_{q_{i-1}}\Pi(x)dx-\frac{q_i-q_{i-1}}{b}\bigg],
\end{equation}

The error $\delta$ is computed across both states and actions, meaning that the total volume of the error between the policies can be thought of as error in $\mathcal{A}$ made for a single state group, which leads to the state approximation error being considered $b_S$ times, akin to conditional probabilities.

The state-action total error can be computed as follows
\begin{equation}
    \delta_{a,s}=\sum_{i=1}^{b_S} \bigg[\int^{q^S_i}_{q^S_{i-1}}\Pi_S(s)ds-\frac{i(q^S_i-q^S_{i-1})}{b_S}\bigg]\sum_{j=1}^{b_A} \bigg[\int^{q^A_{i,j}}_{q^A_{i,j-1}}\Pi_{A}(a)da-\frac{j(q^A_{i,j}-q^A_{i,j-1})}{b_A}\bigg],
\end{equation}
where superscripts and subscripts $A$ and $S$ denotes dependence of the quantity on either action or state, respectively. The notation $q_{i,j}$ refers to the quantile $\hat{\mathcal{Q}}(\frac{i}{b})$ computed conditional on the state falling into bin $i$.
\end{proof}

\begin{proof}(Lemma~\ref{lemma:bandit_reward_approximation})
\begin{equation*}
\begin{split}
    \bigg|\eta(\pi_1)-\eta(\pi_2)\bigg|&= \bigg|\sum_{a\in\cA} r(a)\bigg[\pi_{1}(a)-\pi_{2}(a)\bigg]\bigg|\\
    &= \bigg|\sum_{a\in\cA} r(a)\Delta(a)\bigg|\\
    &\leq \sum_{a\in\cA} \bigg|r(a)\Delta(a)\bigg|\\
    &= ||r\Delta||_1\\
    &\leq ||r||_p||\Delta||_q\\
    &\leq |\cA|^{1/q}M||r||_p\sum_{k=1}^\infty\frac{(\xi_k^{\pi_1}-\xi_k^{\pi_2})^2}{\lambda_{k}}
\end{split}
\end{equation*}
where the before last line is a direct application of Holder inequality and the last line happens because evaluation is a bounded linear functional such that $|f(x)|\leq M||f||_\cH$ for all $f\in\cH$.

\end{proof}

\begin{proof}(Theorem~\ref{thm:mdp_rewards_trpo})
Recall the following result from~\cite{kakade2002approximately,trpo}:
\begin{equation}
    |\eta(\pi_1)-L_{\pi_2}(\pi_1)|\leq \frac{4\alpha^2\gamma \overline{\varepsilon}}{(1-\gamma)^2},
    \label{eq:trpo}
\end{equation}
where $\overline{\epsilon} = \max_{s,a} |Q_{\pi_{1}}(s,a)-V_{\pi_{1}}(s)|$ and $\alpha=\max_{s}TV(\pi_1(\cdot|s),\pi_2(\cdot|s))$.

Instead, suppose that for every fixed state $s\in\cS$, there exists an RKHS of all square-integrable conditional densities of the form $\pi(\cdot|s)$. We examine the difference term between policies in this $\cH_s$. Since $\pi_1,\pi_2 \in \cH_s$, $\delta_{1,2}=\pi_1-\pi_2\in\cH_s$ and hence there exist $M_s>0$ such that $|\delta_{1,2}(a)|\leq M_s ||\delta_{1,2}||_{\cH_s}$ for all $a\in \cA$. Then,
\begin{equation}
    \begin{split}
        \sum_{a\in\cA}|\pi_1(a)-\pi_2(a)| 
        &\leq \sum_{a\in\cA}|\delta_{1,2}(a)| \\
        &\leq M_s \sum_{a\in \cA}||\delta_{1,2}||_{\cH_s}\\
        & \leq |\cA| M_s  \sum_{k=1}^\infty\frac{ (\xi_k^{\pi_1}-\xi_k^{\pi_2})^2}{\lambda_k}\\
        &=|\cA| M_s \Delta_{\cH_s}^\infty,
    \end{split}
    \label{eq:diff_pi_RKHS}
\end{equation}
where we let $\Delta_{\cH_s}^\infty=\sum_{k=1}^\infty\frac{ (\xi_k^{\pi_1}-\xi_k^{\pi_2})^2}{\lambda_k}$ for shorter notation.

Then,
\begin{equation}
    |\eta(\pi_1)-L_{\pi_2}(\pi_1)|\leq \frac{4|\cA|^2\gamma \overline{\varepsilon}}{(1-\gamma)^2}(\displaystyle{\max_{s\in\cS}}M_s \Delta_{\cH_s}^\infty)^2,
\end{equation}

We can obtain a performance bound by first expanding the left hand side
\begin{equation*}
    L_{\pi_2}(\pi_{1})=\eta(\pi_{1})+\sum_{s}\rho_{\pi_{1}}(s)\sum_{a}\pi_{2}(a|s)A_{\pi_{1}}(s,a)
\end{equation*}

Using Eq.~(\ref{eq:trpo}), and using reverse triangle inequality, we have
\begin{equation*}
\begin{split}
 \bigg| \eta(\pi_{2}) -\eta({\pi_{1}}) \bigg|  -  \bigg| \sum_{s}\rho_{\pi}(s)\sum_{a}\pi_{2}(a|s)A_{\pi_{1}}(s,a)  \bigg| &\leq 
 \bigg|\eta(\pi_{2}) -\big\{\eta(\pi_{1})+ \sum_{s}\rho_{\pi}(s)\sum_{a}\pi_{2}(a|s)A_{\pi_{1}}(s,a)  \big\}\bigg|  \\
 &\leq \frac{4\alpha^2\gamma \overline{\varepsilon}}{(1-\gamma)^2}
\end{split}
\end{equation*}

Therefore,
\begin{equation}
\begin{split}
     \bigg| \eta(\pi_{2}) - \eta(\pi_1) \bigg|  &\leq  \frac{4\alpha^2\gamma \overline{\varepsilon}}{(1-\gamma)^2} +\bigg| \sum_{s}\rho_{\pi_{1}}(s)\sum_{a}\pi_{2}(a|s)A_{\pi_{1}}(s,a)  \bigg|  \\
   &\leq  \frac{4\alpha^2\gamma \overline{\varepsilon}}{(1-\gamma)^2}    +   \overline{\epsilon}
\end{split}
\label{eq:eta_diff_trpo}
\end{equation}

We combine Eq.~(\ref{eq:diff_pi_RKHS}) with Eq.~(\ref{eq:eta_diff_trpo}) to obtain the following RKHS form on the improvement lower bound.
\begin{align}
        |\eta(\pi_2)-\eta(\pi_1)| \leq \frac{4 |\cA|^2 \overline{\epsilon} \gamma }{(1-\gamma)^2}(\displaystyle{\max_{s\in\cS}}M_s \Delta_{\cH_s}^\infty)^2 +\overline{\epsilon}
        \label{eq:eta_pi_diff_RKHS}
\end{align}

If we suppose that weights after $K$ are small enough, we can split the approximation error using Lemma~\ref{lemma:rkhs_geom} for $E_K=\frac{\varepsilon^{2(K+1)}}{1-\varepsilon^2}$:
\begin{equation}
\begin{split}
    M_s\Delta_{\cH_s}^\infty \leq M_s \bigg(\Delta_{\cH_s}^K+E_K\bigg),\;.
\end{split}
\end{equation}

Using this in Eq.~(\ref{eq:eta_pi_diff_RKHS}) yields the final result:
\begin{align}
    |\eta(\pi_2)-\eta(\pi_1)| &\leq \frac{4 |\cA|^2 \overline{\epsilon} \gamma }{(1-\gamma)^2}\bigg(\displaystyle{\max_{s\in\cS}}(M_s \Delta_{\cH_s}^K)^2+2E_K\displaystyle{\max_{s\in\cS}}M_s \Delta_{\cH_s}^K+E_K^2\max_{s\in\cS}M_s^2\bigg) +\overline{\epsilon} \\
     |\eta(\pi_2)-\eta(\pi_1)| &\leq \frac{4 |\cA|^2 \overline{\epsilon} \gamma }{(1-\gamma)^2}\bigg\{\displaystyle{\max_{s\in\cS}}(M_s \Delta_{\cH_s}^K)^2+O(\varepsilon^{2K}\displaystyle{\max_{s\in\cS}}M_s \Delta_{\cH_s}^K)\bigg\} +\overline{\epsilon}\;.
 \end{align}.
 
Now, if $\varepsilon$ is close to zero (i.e. $\pi_1$ and $\pi_2$ have very similar $\xi_K,\xi_{K+1}..$), then the expression simplifies to
\begin{equation}
    |\eta(\pi_2)-\eta(\pi_1)| \leq \frac{4 |\cA|^2 \overline{\epsilon} \gamma }{(1-\gamma)^2}\displaystyle{\max_{s\in\cS}}(M_s \Delta_{\cH_s}^K)^2 +\overline{\epsilon}\;.
\end{equation}

\end{proof}

\begin{proof}(Theorem~\ref{thm:rho_minus_rho_tilde})
We first represent the approximate transition matrix $\mat{P}_{\tilde{\pi}}$ induced by the policy $\tilde{\pi}$ as a perturbation of the true transition:
\begin{equation}
\begin{split}
     \mat{P}_{\tilde{\pi}} &= \mat{P}_\pi + (\mat{P}_{\tilde{\pi}}-\mat{P}_\pi)\\
     &= \mat{P}_\pi + \mat{E}
\end{split}
\end{equation}
Then, the difference between stationary distributions $\rho_{\tilde{\pi}}$ and $\rho_\pi$ is equal to~\cite{schweitzer1968perturbation,cho2001comparison}:
\begin{equation}
    \rho_\pi^\top-\rho_{\tilde{\pi}}^\top = \rho_{\tilde{\pi}}^\top\mat{E}\mat{Z},
\end{equation}
where $\mat{Z}=(\mat{I}-\mat{P}_\pi+\mat{1}_{|\cS|}\rho_\pi^\top)^{-1}$ is the \textit{fundamental matrix} of the Markov chain induced by $\mat{P}_\pi$ and $\mat{1}_{|\cS|}$ is a vector of ones.

In particular, the above result holds for Schatten norms~\cite{baumgartner2011inequality}:
\begin{equation}
    ||\rho_\pi-\rho_{\tilde{\rho}}||_{2} \leq ||\rho_{\tilde{\pi}}||_{2} ||\mat{Z}||_{S_\infty} ||\mat{E}||_{S_\infty}\leq  ||\mat{Z}||_{S_\infty} ||\mat{E}||_{S_1}
\end{equation}

So far, this result is known for irreducible, homogeneous Markov chains and has no decision component.\\
Consider the matrix $\mat{E}$, which is the difference between expected transition models of true and approximate policies. It can be expanded into products of matricized tensors:
\begin{equation}
    \begin{split}
        \mat{E} &= \mat{P}_{\tilde{\pi}}-\mat{P}_\pi\\
        &= \mat{A}(\tilde{\mat{\Pi}}\otimes \mat{I})\mat{T}_{(3)}-\mat{A}(\mat{\Pi}\otimes \mat{I})\mat{T}_{(3)}^\top\\
        &= \mat{A}((\tilde{\mat{\Pi}}-\mat{\Pi} )\otimes \mat{I})\mat{T}_{(3)}^\top
    \end{split}
\end{equation}

The norm of $\mat{E}$ can also be upper bounded as follows:
\begin{equation}
\begin{split}
    ||\mat{E}||_{S_1} &\leq ||\mat{A}||_{S_1}||(\tilde{\mat{\Pi}}-\mat{\Pi} )\otimes \mat{I}\mat{T}_{(3)}^\top||_{S_\infty}\\
    &\leq ||\mat{A}||_{S_1}||(\tilde{\mat{\Pi}}-\mat{\Pi} )\otimes \mat{I}\mat{T}_{(3)}^\top||_{S_1}\\
    &\leq ||\mat{A}||_{S_1}||(\tilde{\mat{\Pi}}-\mat{\Pi} )\otimes \mat{I}||_{S_2}||\mat{T}_{(3)}^\top||_{S_2}\\
    &\leq ||\mat{A}||_{S_1}||(\tilde{\mat{\Pi}}-\mat{\Pi} )\otimes \mat{I}||_{S_1}||\mat{T}_{(3)}^\top||_{S_2}\\
    &\leq ||\mat{A}||_{S_1}||\tilde{\mat{\Pi}}-\mat{\Pi}||_{S_1} ||\mat{I}||_{S_\infty}||\mat{T}_{(3)}^\top||_{S_2}\\
    &= ||\tilde{\mat{\Pi}}-\mat{\Pi}||_{S_1} ||\mat{T}_{(3)}||_{S_2}\\
\end{split}
\end{equation}

Combining this result from that of~\cite{schweitzer1968perturbation} yields
\begin{equation}
    ||\rho_\pi-\rho_{\tilde{\rho}}||_{S_1} \leq \underbrace{ ||\mat{Z}||_{S_\infty} }_{ \substack{\text{Depends on}\\ \text{MDP + $\pi$}} } \underbrace{||\tilde{\mat{\Pi}}-\mat{\Pi}||_{S_1}}_{ \substack{\text{Depends on $\pi$}} } \underbrace{||\mat{T}_{(3)}||_{S_2}}_{ \substack{\text{Depends on MDP}} }
\end{equation}
\end{proof}

\subsection{Time complexity comparison for different approximation methods}
\label{time_complexity}

The time complexity for a $k$-truncated SVD decomposition of a matrix $A\in \mathbb{R}^{n\times m}$ is $O(mnk)$ as discussed in~\cite{du2017power}. Fast Fourier transform can be done in $O(mn\log(mn))$ for the 2-dimensional case~\cite{lohne2017computational}. Hence, SVD is expected to be faster whenever $k<\log(nm)$. The discrete wavelet transform's (DWT) time complexity depends on the choice of mother wavelets. When using Mallat's algorithm, the 2-dimensional DWT is known to have complexities ranging from $O(n^2\log n)$ to as low as $O(n^2)$, for a square matrix of size $n \times n$ and depending on the choice of mother wavelet~\cite{resnikoff2012wavelet}.
Finally, we expect FFT and Daubechies wavelets to have less parameters than SVD, since the former use pre-defined orthonormal bases, while the later must store the left and right singular vectors.

\subsection{Rate of convergence of stationary distribution under random policy}

\begin{figure}[h!]
    \centering
    \includegraphics[width=\linewidth]{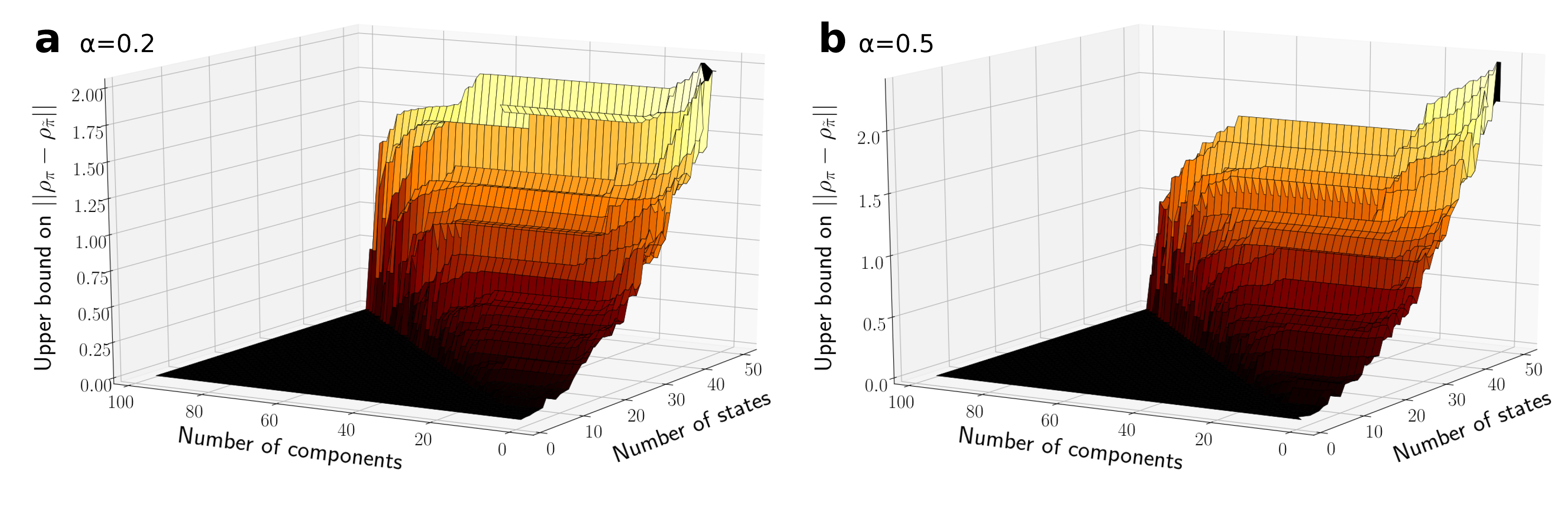}
    \caption{Upper bound on $||\rho_\pi-\rho_{\hat{\pi}}||$ in the chain MDP task as a function of number of Fourier components and of the number of states.}
    \label{fig:rho_minus_rho_tilde}
\end{figure}

We validate the rate of convergence of Thm.~\ref{thm:rho_minus_rho_tilde} in the toy experiment described below. Consider a deterministic chain MDP with $N$ states. A fixed policy in state $i$ transitions to $i+1$ with probability $\alpha$ and to $i-1$ with probability $1-\alpha$. If in states $1$ or $N$, the agent remains there with probability $1-\alpha$ and $\alpha$, respectively. We consider the expected transition model $\mat{P}_\pi$ obtained using the policy reconstructed through discrete Fourier transform. A visualization of Theorem~\ref{thm:rho_minus_rho_tilde} is shown in Fig.~\ref{fig:rho_minus_rho_tilde}.

\subsection{Experiment details}
This subsection covers all aspects of the conducted experiments, including pseudocode, additional results and details about the setup.
\subsubsection{Python pseudocode}
\label{sec:python_code}
Below is the Python-like pseudocode for the case of Fourier bases (which can be easily adapted to all bases examined in this paper). This pseudocode is heavily simplified Python, and skips some bookkeeping steps such as edge cases, but convey the overall flow of the algorithm. 
\input{code_snippet}
\subsubsection{Bandit turntable}
We consider a bandit problem in which rewards are spread in a circle as shown in Figure~\ref{fig:bandit_turntable} a). Actions consist of angles in the interval $ [-\pi,\pi]$. Given the corresponding Boltzmann policy, we compare the reconstruction quality of the discrete Fourier transform and GMM. In addition to Fourier and GMM, we also compare with fixed basis GMM, where the means are sampled uniformly over $ [-\pi,\pi]$ and variance is drawn uniformly from $\{0.1,0.5,1,2\}$; only mixing coefficients are learned. This provides insight into whether the reconstruction algorithms are learning the policy, or if $\pi$ is so simple that a random set of Gaussian can already approximate it well.

\subsubsection{Experimental parameters} \label{app:exp_parameters}
All parameters were chosen using a memory-performance trade-off heuristic. For example, for Continuous Mountain Car, we based ourselves off Figure~\ref{fig:cmc_large_fig} to select $b_S=35$.
\begin{table}[h!]
    \centering
\begin{tabular}{c|c |c| c}
\toprule
Hyperparameters    & Turntable & Pendulum & Continuous Mountain Car  \\ 
\midrule
$b_A$   & 100     & 15 &  10  \\
$b_S$              & N/A    & 35 &     35  \\ 
\hline
Optimizer      & \multicolumn{3}{c}{Adam}   \\ 
Architecture      & \multicolumn{3}{c}{256}   \\ 
Learning rate      & \multicolumn{3}{c}{1e-03}   \\ 
Hidden dimension        & \multicolumn{3}{c}{256}    \\
$\#$ rollouts      & \multicolumn{3}{c}{100}   \\ 
Torch seeds       & \multicolumn{3}{c}{0 to 9} \\   
\bottomrule
\end{tabular}
    \caption{Hyperparameters used across experiments, N/A: not applicable}
\end{table}

\subsubsection{Pendulum}
 This classical mechanic task consists of a pendulum which needs to swing up in order to stay upright. The actions represent the joint effort (torque) between $-2$ and $2$ units. We train SAC until convergence and save snapshots of the actor and critic after $2k$ and $5k$ steps. The reconstruction task is to recover both the Gaussian policy (actor) and the Boltzmann-Q policy (critic, temperature set to 1). 
 \begin{figure}[H]
    \centering
    \includegraphics[width=0.5\linewidth]{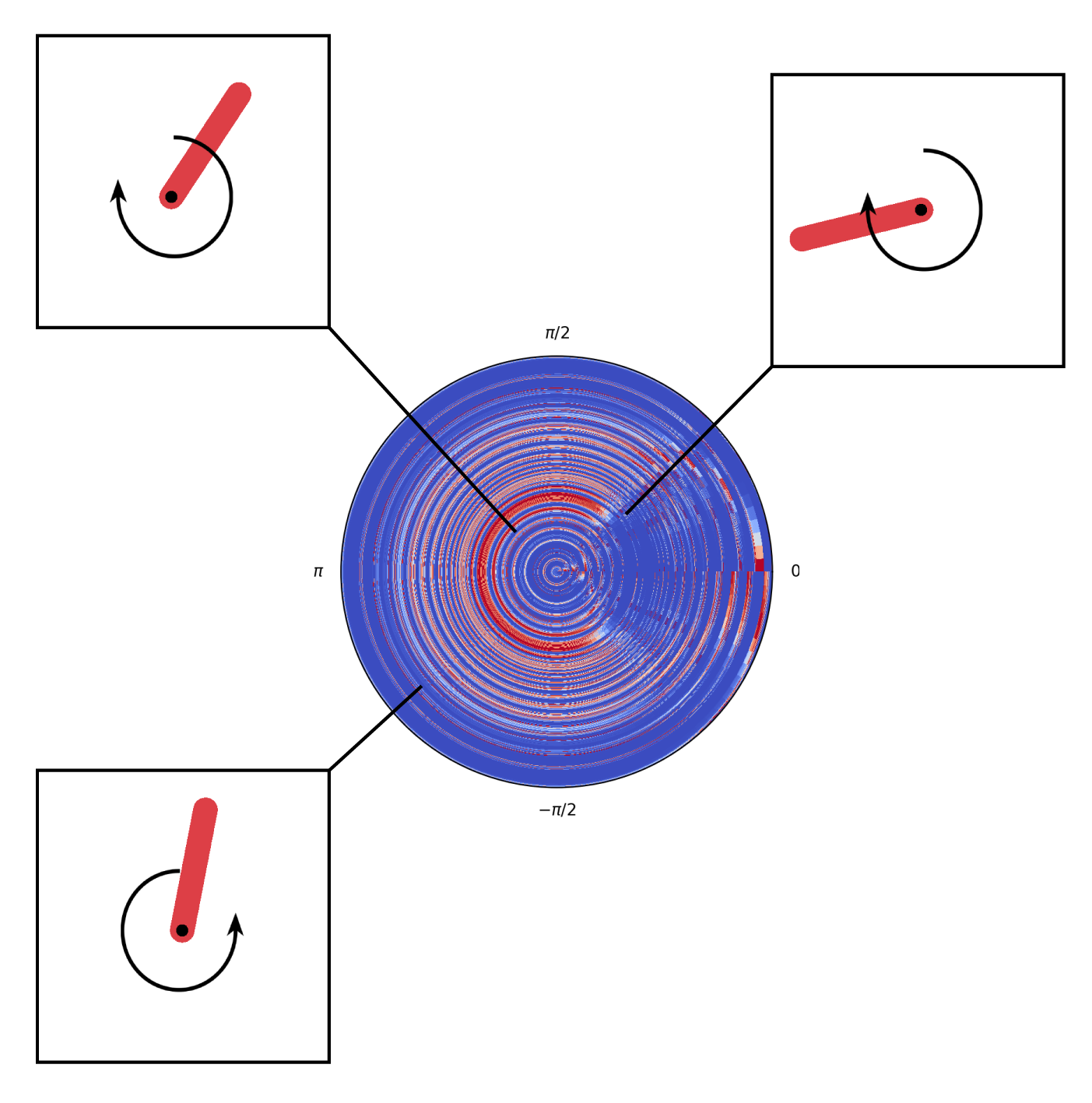}
    \caption{Plots of Gaussian $\Pi$ after $5k$ training steps in polar coordinates. Each circle of different radius represents a states in the \texttt{Pendulum} environment as shown by the snapshots. Higher intensity colors (red) represent higher density mass on the given angular action.}
    \label{fig:cartoon_pendulum}
\end{figure}

\begin{figure}[H]
    \centering
    \includegraphics[width=0.7\linewidth]{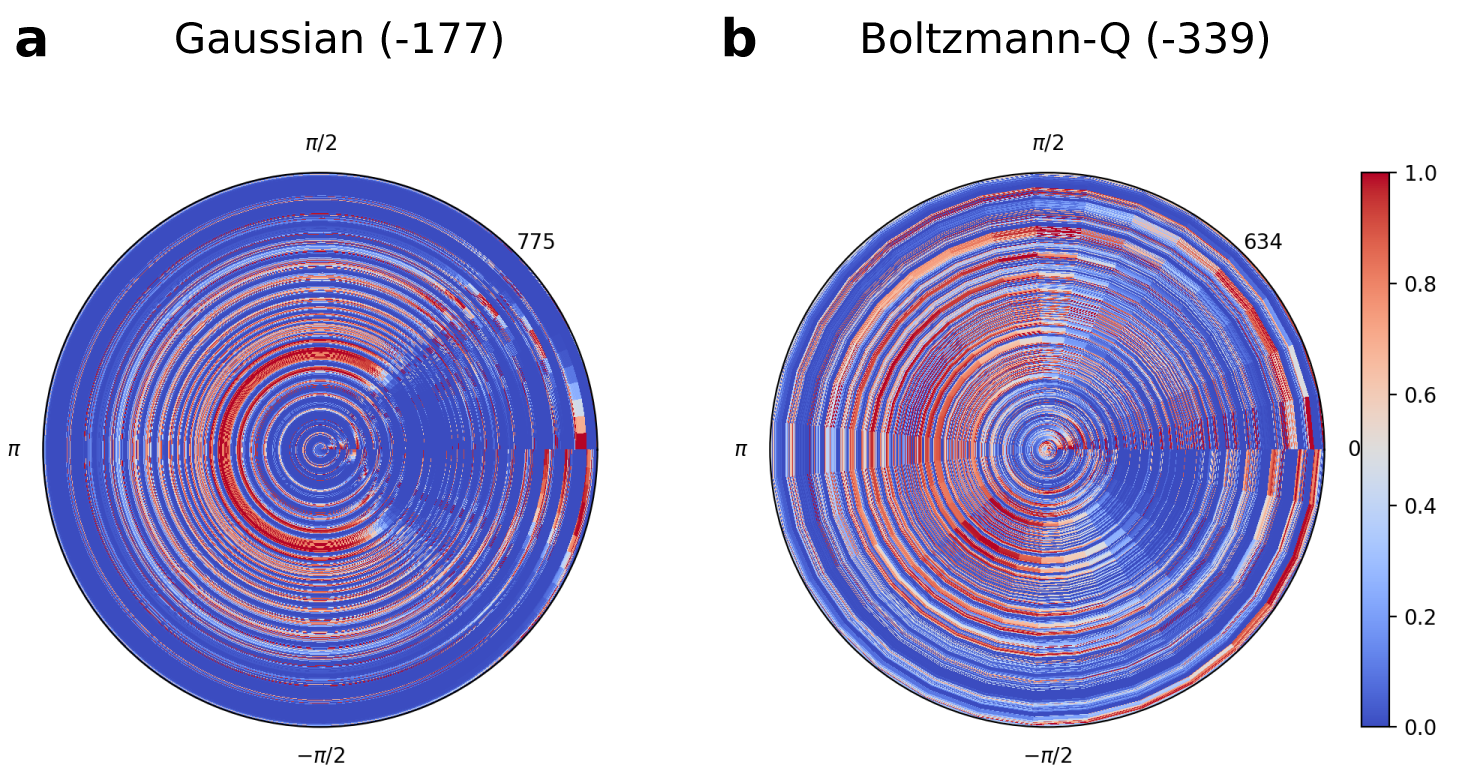}
    \caption{Plots of discretized and pruned (a) Gaussian policy and (b) Boltzmann-Q policy in polar coordinates for $5$ thousands training steps of soft actor-critic. Rewards collected by their respective continuous networks are indicated in parentheses. Each circle of radius $s$ corresponds to $\pi(\cdot|s)$ for a discrete $s=1,2,..,750$. All densities are on the same scale.}
    \label{fig:radial_gaussian_pendulum}
\end{figure}


\begin{figure}[H]
    \centering
    \includegraphics[width=0.7\linewidth]{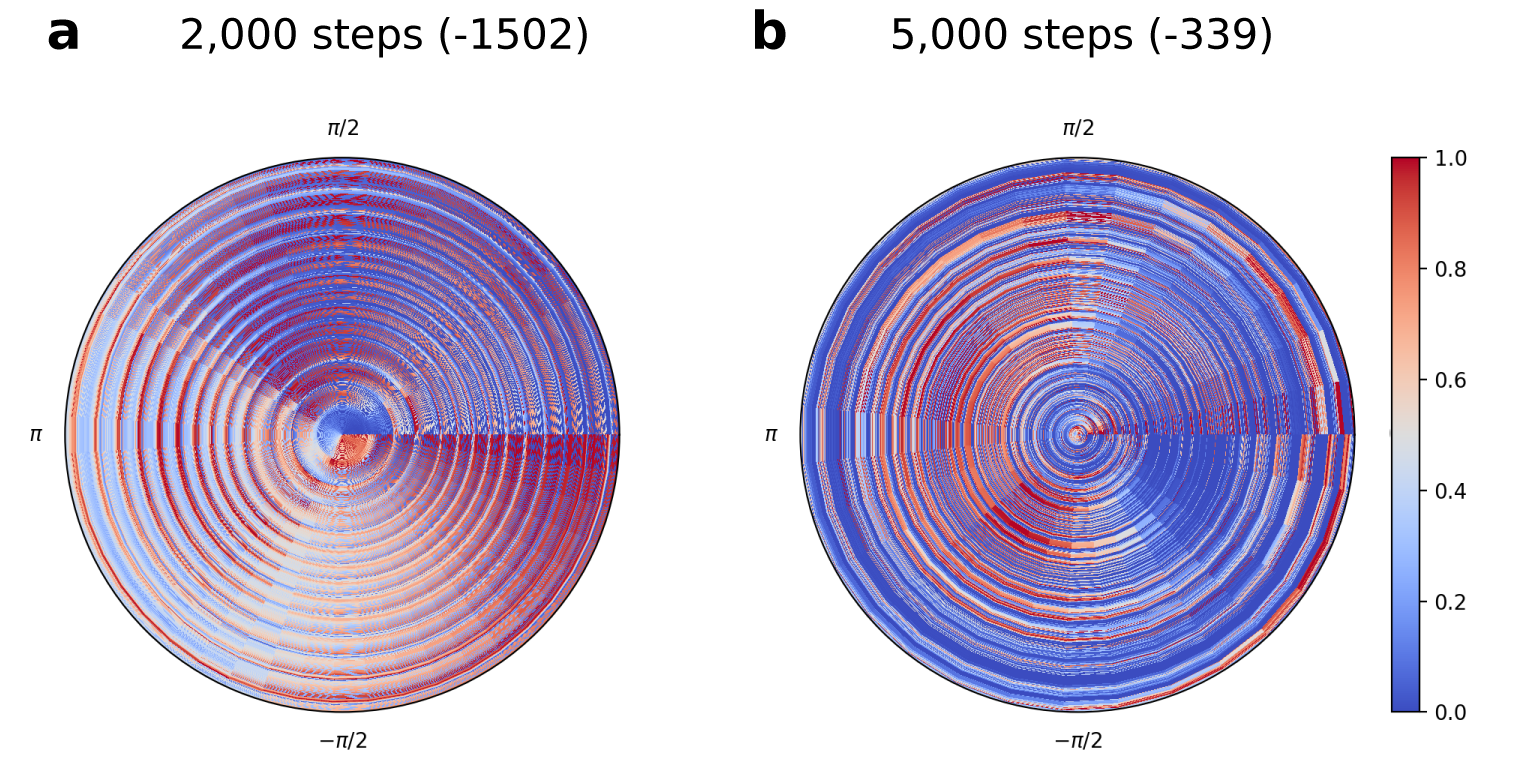}
    \caption{Plots of discretized and pruned Boltzmann policy in polar coordinates for $2$ and $5$ thousands training steps of soft actor-critic. Each circle of radius $s$ corresponds to $\pi(\cdot|s)$ for a discrete $s=1,2,..,750$.}
    \label{fig:radial_boltzmann_pendulum}
\end{figure}

\begin{figure}[H]
    \centering
    \includegraphics[width=0.7\linewidth]{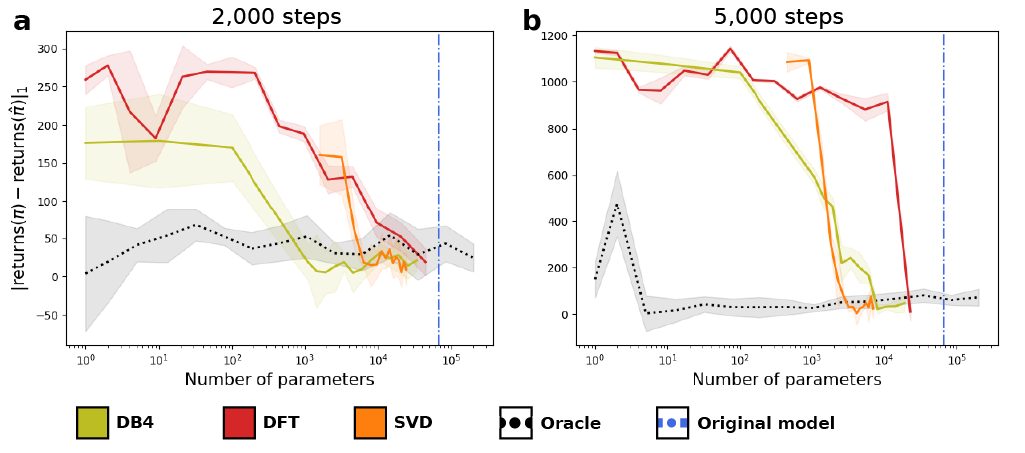}
    \caption{Absolute difference in returns collected by discretized and reconstructed Boltzmann-Q policies for \textit{Pendulum-v0}, averaged over 10 trials. Blue dots represent the number of parameters of the neural network policy. SVD, DFT and DB4 projections need an order of magnitude less in term of parameters to reconstruct the original policy.}
    \label{fig:rewards_pendulum_boltzmann}
\end{figure}

\begin{figure}[H]
    \centering
    \includegraphics[width=0.7\linewidth]{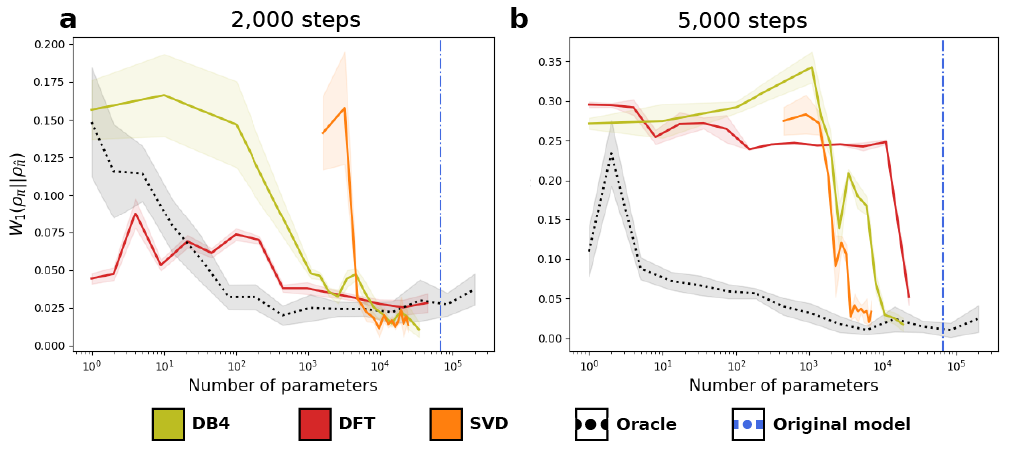}
    \caption{$W_1$ distance between the true and reconstructed stationary Boltzmann-Q distributions, averaged over 10 trials for \textit{Pendulum-v0}.}
    \label{fig:W1_distance_rho_pi_boltzmann_pendulum}
\end{figure}


\begin{figure}[h!]
    \centering
    \includegraphics[width=0.7\linewidth]{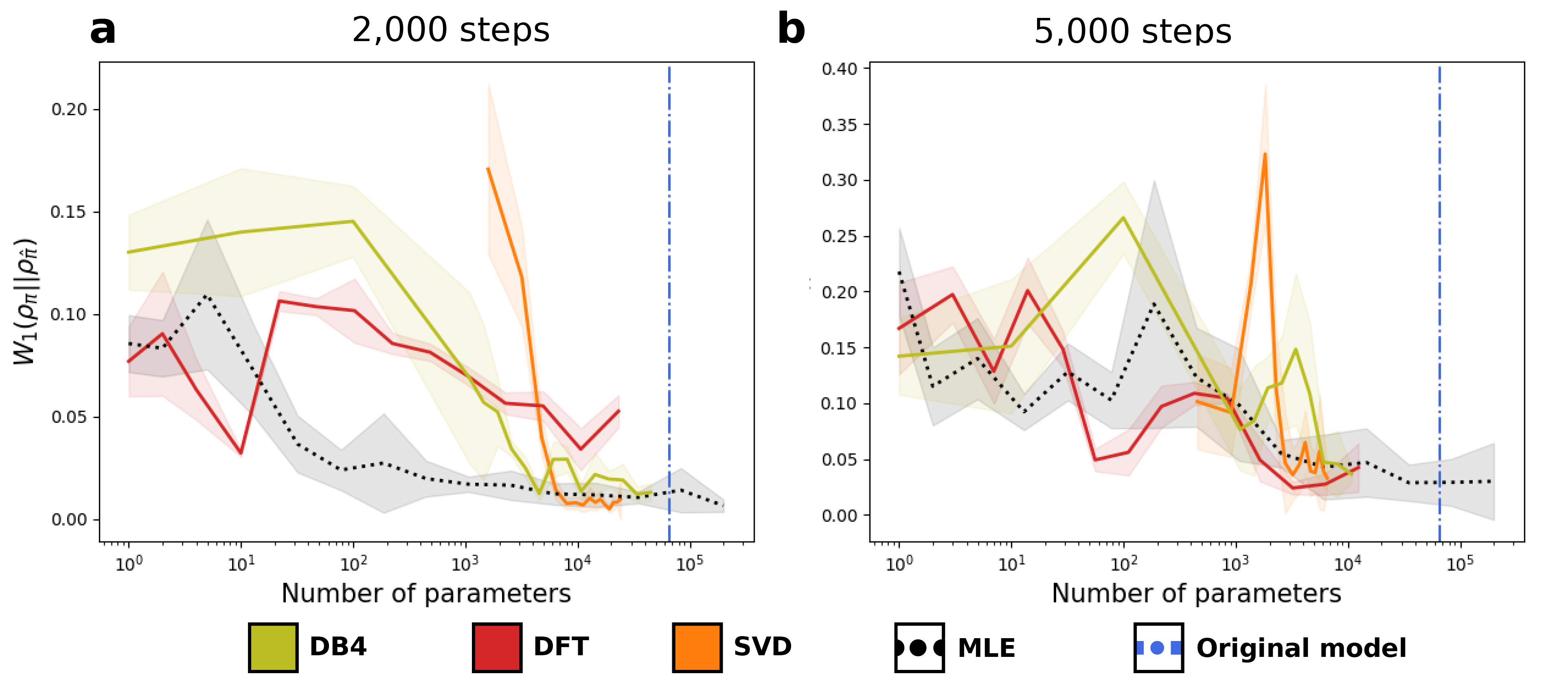}
    \caption{$W_1$ distance between the true and reconstructed stationary distributions, averaged over 10 trials. SVD, DFT and DB4 methods show a fast convergence to the oracle's stationary distribution.}
    \label{fig:W1_distance_rho_pi}
\end{figure}

\subsubsection{Mountain Car}
In this environment, the agent (a car) must reach the flag located at the top of a hill. It needs to go back and forth in order to generate sufficient momentum to reach the top. The agent is allowed to apply a speed motion in the interval $[-1,1]$.
We use Soft Actor-Critic (SAC, \cite{sac}) to train the agent until convergence~($25k$ steps). 

\begin{figure}
    \centering
    \includegraphics[width=\linewidth]{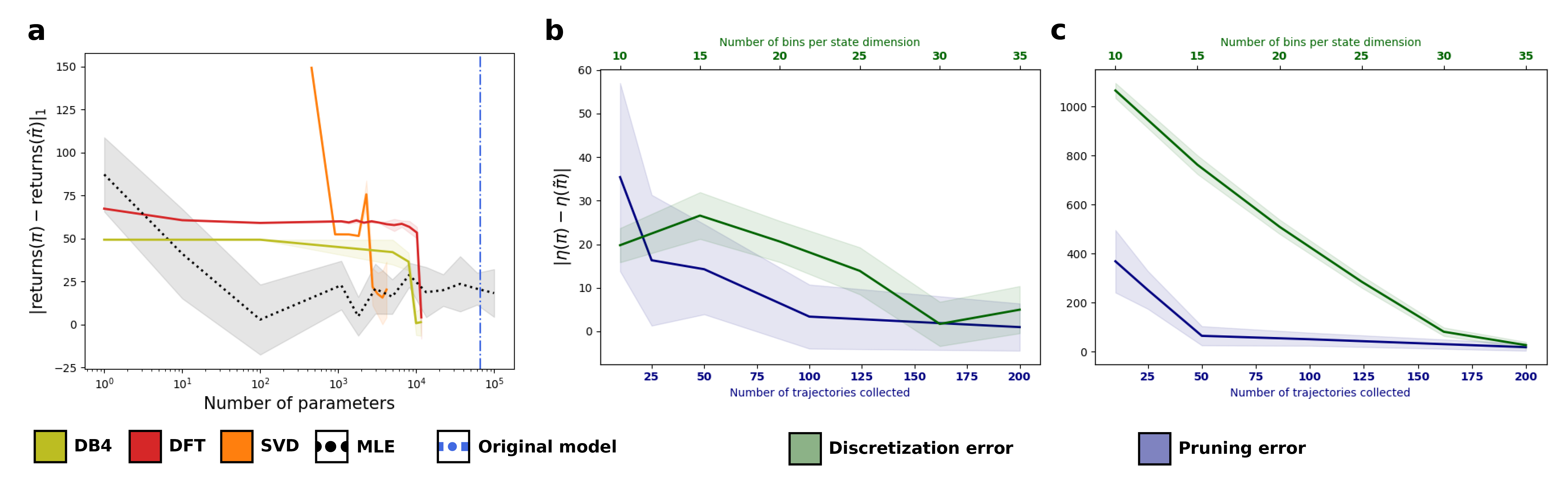}
    \caption{\textbf{(a)} Difference in returns between embedded and ground truth policies for the Mountain Car task with respect of the number of parameters used and the discretization and pruning errors for \textbf{(b)} the Gaussian policy of \texttt{ContinuousMountainCar-v0} and \textbf{(c)} the Boltzmann-Q policy of \texttt{Pendulum-v0}}
    \label{fig:cmc_large_fig}
\end{figure}

Figure~\ref{fig:cmc_large_fig}a shows that the error between the true and truncated policies steadily decreases as a function of projection weights kept. Note that SVD's parameters are computed as total entrie in the matrices $U,D,V$, and hence the smallest possible rank (rank 1) of $D$ dictates the minimal number of parameters. Figure~\ref{fig:cmc_large_fig}b-c show how the discretization and prunning performances behave as a function of trajectories (for pruning, since it relies on trajectories to estimate $\rho_\pi$), or bins per state dimension (for discretization).

\subsection{Usefulness of the pruning step}
\label{sec:pruning_motivation}
In an environment where the state space is very large, or when policies are degenerate along a narrow path in the state space, the pruning step can turn out to be very useful.

Table~\ref{tab:pruning_percentage} shows the proportion of parameters from the discretized policy that was pruned. Pruned states are ones which are visited with probability at most $\varepsilon$; in all our experiments, $\varepsilon$ was set to 0, so no visited state was ever pruned.

\begin{table}[h!]
    \centering
    \begin{tabular}{c|c}
    \toprule
        Task & Pruned \% \\
        \midrule
        Pendulum (2K) & 95.36\\
        Pendulum (3K) & 94.49\\
        Pendulum (4K) & 94.59\\
        Pendulum (5K) & 98.45\\
        \bottomrule
    \end{tabular}
    \caption{Percentage of pruned parameters for the Pendulum environment policies}
    \label{tab:pruning_percentage}
\end{table}

The pruning step is extremely efficient in the Pendulum task, allowing to reduce the memory complexity between the discretization and the truncation step by up to 98\%.

Note that as policies get trained on more samples, they concentrate along the optimal policy. This in turn means that less exploration is needed, and hence the policy coverage gets reduced, allowing us to prune more states.

%% file: code_snippet.tex
\begin{minted}[mathescape,
               linenos,
               numbersep=5pt,
               gobble=2,
               frame=lines,
               framesep=2mm]{python}
               
    def RKHS_policy_step_1_and_2(cts_pi,env,n_trajectories,n_episode_steps,b_S,b_A):
        """
        cts_pi: Object with 2 methods: 
            - sample(state): samples an action from pi(.|state);
            - log_prob(state,action): returns pi(a|s);
        env: OpenAI Gym environment;
        n_trajectories: Number of trajectories in rollouts (positive integer);
        n_episode_steps: Number of maximum steps in each episode (positive integer);
        b_S: number of state bins per dimension (positive integer);
        b_A: number of action bins per dimension (positive integer).
        """
        from sklearn.preprocessing import KBinsDiscretier
        states,actions,rewards = rollout(cts_pi,env.n_trajectories,n_episode_steps)
        # We use quantile in the paper but KMeans can also be used to find the bins
        enc_A = KBinsDiscretizer(n_bins=b_A,encode='ordinal',strategy='quantile')
        enc_S = KBinsDiscretizer(n_bins=b_S,encode='ordinal',strategy='quantile')
        bins_A = enc_A.fit_transform(actions)
        bins_S = enc_S.fit_transform(states)
        
        dsc_pi = np.zeros(shape=(b_S,b_A))
        for i in range(b_S):
            for j in range(b_A):
                s,a = enc_S.inverse_transform([i]), enc_A.inverse_transform([j])
                dsc_pi[i,j] = np.exp(cts_pi.log_prob(s,a))
            dsc_pi[i,:] /= np.sum(dsc_pi[i,:) # re-normalize every conditional pmf
        
        rho_pi = np.zeros(shape=(b_S,))
        for bin in bins_S:
            rho_pi[bin] += 1
        rho_pi /= np.sum(rho_pi)
        
        return dsc_pi, rho_pi
\end{minted}

\begin{minted}[mathescape,
               linenos,
               numbersep=5pt,
               gobble=2,
               frame=lines,
               framesep=2mm]{python}
               
    def RKHS_policy_step_3(dsc_pi,rho_pi):
        """
        dsc_pi: np.array of size b_S x b_A;
        rho_pi: np.array of length b_S.
        """
        idx_to_keep = np.where(rho_pi>0)
        
        pruned_dsc_pi = dsc_pi[idx_to_keep]
        
        def pruned_agent(state_bin):
            actions = np.arange(0,b_A)
                if state_bin in idx_to_keep:
                    probs = pruned_dsc_pi[idx_to_keep==state_bin] # probabilities from the lattice
                else:
                    probs = actions*0+1/actions.size # uniform probabilities
                return np.random.choice(actions,1,p=probs)
        
        return pruned_agent
\end{minted}

\begin{minted}[mathescape,
               linenos,
               numbersep=5pt,
               gobble=2,
               frame=lines,
               framesep=2mm]{python}
               
    def RKHS_policy_step_4(dsc_pi,K):
        """
        dsc_pi: np.array of size b_S x b_A;
        K: number of components to keep (positive integer).
        """
        FS = np.fft.fftn(dsc_pi) # transform to Fourier domain
        fshift = np.real(np.fft.fftshift(FS)).flatten() # vector of size b_S x b_A
        # find threshold s.t. exactly K components are not zeroed out
        threshold = sorted(flat_fshift,reverse=True)[K]
        mask = np.real(fshift >= th).astype(int) 
        fshift *= mask # apply the binary mask to zero out \xi_K,\xi_K+1,...
        f_ishift = np.fft.ifftshift(fshift) # inverse shift of truncated spectrum
        f_dft = np.real(np.fft.ifftn(f_ishift))
\end{minted}